\documentclass[acp, manuscript]{copernicus}

\newcommand{\ncufp}{\ensuremath{\mathrm{NC}_{\mathrm{UFP}}}}
\newcommand{\nctot}{\ensuremath{\mathrm{NC}_{\mathrm{tot}}}}
\graphicspath{{figures/}}
\usepackage{placeins}

\begin{document}
\nolinenumbers 

\title{Estimating aerosol particle number size distributions from PM$_{2.5}$ with physics-regularized flow matching}

\Author[1,2,3][zhisong.liu@lut.fi]{Zhi-Song}{Liu}
\Author[1,2,3]{Jenni}{Köykkä}
\Author[2,3]{Wenqing}{Peng}
\Author[1,2,3]{Ammar}{Kheder}
\Author[2,3]{Petri}{Clusius}
\Author[1,2,3]{Michael}{Boy}

\affil[1]{Department of Computational Engineering, LUT University, Finland} \affil[2]{Atmospheric Modelling Center, Lahti (AMC-Lahti), Finland} \affil[3]{Institute for Atmospheric and Earth System Research (INAR), University of Helsinki, Finland}

\runningtitle{Probabilistic estimation of particle number size distributions}
\runningauthor{Z.-S. Liu et al.}

\received{}
\pubdiscuss{}
\revised{}
\accepted{}
\published{}

\firstpage{1}

\maketitle

\begin{abstract}
The mass concentration of particles smaller than 2.5\,$\mu$m (PM$_{2.5}$) is cheap to measure and widely monitored, whereas the particle number size distribution (PNSD) requires costly instruments and is measured at only a few research stations. Yet many aerosol effects, such as vapor uptake, cloud droplet formation and deposition in the human respiratory tract, depend on the PNSD rather than on mass. We explore whether generative deep learning models can infer the current PNSD from routine data: from the PM$_{2.5}$ record and eight meteorological variables over a 24\,h window, a model generates an ensemble of plausible hourly PNSDs for the same window, reflecting that one mass can arise from many size distributions. Rather than proposing a single best model, we use two contrasting Finnish stations, the boreal forest site Hyyti\"al\"a and the urban site Kumpula, as a controlled testbed. On it, we benchmark six generative models against deterministic regression and historical-mean baselines, locally and after transfer between sites, and test physical constraints for neural network training. Flow matching performs best overall; its guided variant has the lowest continuous ranked probability score (CRPS) at both sites. A constraint based on a reduced aerosol general dynamic equation (GDE) lowers the CRPS by 5.5--11.5\,\% at the training site and reduces errors after transfer between sites. Transfer remains harder than local estimation, identifying multi-station training as the next step towards PNSD estimates at unmonitored locations. This benchmark establishes physics-informed generative models as a promising route from routine air-quality monitoring to size-resolved aerosol information.
\end{abstract}

\introduction

Particle number and particle mass describe different aspects of an aerosol population. Newly formed or freshly emitted ultrafine particles can dominate the total number concentration while contributing negligibly to mass, whereas far fewer accumulation-mode particles can dominate PM$_{2.5}$, the mass concentration of particles smaller than 2.5\,$\mu$m in aerodynamic diameter. The particle number size distribution (PNSD) is the state on which aerosol processes act: condensational growth, coagulation and deposition change it, and its shape in turn determines how much and which vapors condense onto the particles, how many particles can activate to cloud droplets and where they deposit in the respiratory tract \citep{seinfeld2016,kulmala2001}. We use the term PNSD throughout, rather than the more general ``size distribution'', to make clear that it refers to the number of particles per logarithmic diameter interval, $\mathrm dN/\mathrm d\log_{10}D_p$, and not to a mass or surface-area distribution. Air-quality networks routinely report PM$_{2.5}$, whereas measuring the PNSD requires dedicated instruments, such as differential or scanning mobility particle sizers (DMPS, SMPS) and, for the smallest particles, ion spectrometers and particle size magnifiers, all of which need specialist operation and quality control \citep{kulmala2012,wiedensohler2012,hartl2026}. The practical question is therefore how much of the PNSD can be recovered from PM$_{2.5}$ and meteorological data.

Two families of methods approach this question from opposite directions. Process-based models, ranging from global Earth system models and regional chemical transport models such as WRF-Chem \citep{grell2005}, Enviro-HIRLAM \citep{baklanov2017} and SILAM \citep{sofiev2015} to Lagrangian models such as SOSAA-FP \citep{boy2011,clusius2026} and ADCHEM \citep{roldin2019}, solve the \emph{forward} problem: given emissions, meteorology and chemistry, they simulate how the PNSD evolves by resolving gas-phase chemistry, boundary-layer dynamics and aerosol microphysics. Their inputs, however, include emission inventories, vertical meteorology and gas-phase concentrations that are not normally available at a monitoring station. Recovering the PNSD from PM$_{2.5}$ is instead an \emph{inverse} problem: the size distribution must be inferred from a single mass integral of it. Because this inverse problem has many solutions, other local observables, such as meteorology, can be added as constraints. Machine-learning methods suit such problems, since they learn the mapping from an input space to an output space directly from data. Deep neural networks already estimate PM$_{2.5}$ at locations without measurements \citep{kheder2025} and forecast it at high resolution when physical knowledge, such as the effects of topography and wind, is built into their design \citep{kheder2026}. \citet{vartiainen2025} downscaled climate-model aerosol output to daily number concentrations in three size ranges, and \citet{barahona2026} inferred modal size distributions and mixing state from bulk aerosol mass and meteorology. Together, these works establish that bulk properties contain recoverable size information.

However, both studies return a single deterministic estimate, which cannot represent the central difficulty of this inverse problem: the same PM$_{2.5}$ can arise from a few large particles or many small ones. We refer to this as the \emph{mass--number ambiguity}. A deterministic regression model trained on squared error returns, for each input, the average of all PNSDs compatible with it. This average smooths out the structure that matters for number-sensitive applications, both between neighbouring diameter bins and from hour to hour within a 24\,h window. As shown in Sect.~\ref{sec:res_interpolation}, a gradient-boosting regression reproduces only 58\,\% of the observed variability of the PNSD from one 24\,h window to the next, against 83\,\% for the generative model introduced below. Two further questions are rarely tested. First, a score averaged over all diameter bins can hide errors in the size range on which a particular application depends. We therefore evaluate two quantities derived from the PNSD with contrasting size weighting. The number concentration of ultrafine particles, those smaller than 100\,nm (\ncufp), is dominated by the smallest particles and is relevant for health and new particle formation \citep{who2021}. The condensation sink (CS), the rate at which condensable vapors are lost to the existing particles, is weighted towards accumulation-mode particles \citep{kulmala2001}. Both follow from the same PNSD, but a model can estimate one of them well and the other poorly. Second, a model fitted at one station may not transfer to a station with a different aerosol regime.

\begin{figure*}[t]
\includegraphics[width=0.9\textwidth]{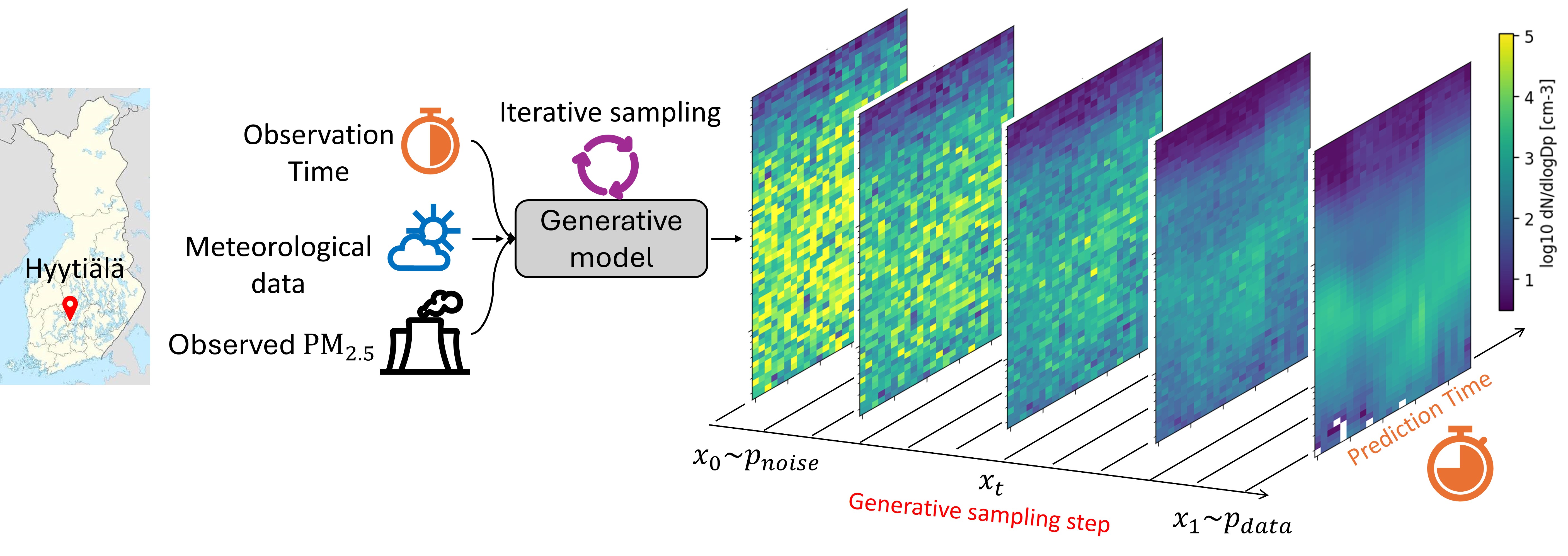}
\caption{\textbf{PNSD estimation as conditional generation.} The same 24\,h of PM$_{2.5}$ and meteorology that condition the transformation of Gaussian noise also cover the generated hourly PNSD: the target window is concurrent with the input record. Repeated noise draws give an ensemble for a fixed input. Generative integration time is distinct from the physical (clock) time of the measurements.}
\label{fig:teaser}
\end{figure*}

Conditional generative models \citep{lipman2023,liu2023,lfm,fm_language,ZSLiu_2025} learn a distribution of outputs for a fixed input, and therefore offer a direct way to represent this ambiguity. Among them, flow matching (FM) learns a smooth transport from Gaussian noise to the data by regressing a velocity field, which makes training a simple regression problem and sampling a fixed number of integration steps \citep{lipman2023,liu2023}. In this work, we use FM to generate an ensemble of hourly PNSDs for a 24\,h window from the concurrent PM$_{2.5}$ and meteorological record (Fig.~\ref{fig:teaser}). We call this task \emph{PNSD estimation}. It is not a forecast: the PNSD is never measured, and the model infers it from other variables observed in the same window, receiving no PNSD measurement at any time. Such an estimate could fill gaps in existing PNSD records or provide a PNSD at stations that measure only PM$_{2.5}$. It also changes the spatial scale of the available information: a PNSD measurement describes the aerosol at a single research station, whereas a computational estimate can be produced wherever PM$_{2.5}$ is monitored, for example at every station of a municipal air-quality network, and thus at a much higher spatial density.

Since generative models have not yet been applied to this problem, our aim is exploratory. Rather than proposing a single best model, we benchmark the main families of generative models for PNSD estimation and test whether physical constraints can guide neural network training. Because a generative objective alone does not enforce aerosol physics, we add soft penalties of two kinds: \emph{integral constraints} on the resolved mass and total number, and a \emph{dynamical constraint} based on a reduced general dynamic equation (GDE). A benchmark of this kind needs controlled conditions more than scale: every model must see the same inputs, training period and evaluation windows, so that differences can be attributed to the modelling choices. We therefore use two well-characterized stations with long, continuous PNSD records as a testbed: the Hyyti\"al\"a boreal forest station and the Kumpula urban station in Finland. Their contrasting aerosol regimes make the pair a deliberately demanding test of transfer between environments. PM$_{2.5}$ and meteorology alone carry little information that distinguishes urban from rural aerosol. Tracers of combustion, such as nitrogen oxides (NO$_x$), would help, but we deliberately restrict the inputs to PM$_{2.5}$ and meteorology, which are available almost everywhere, so that transfer between the two stations is a genuine out-of-distribution test. Within this testbed we address four research questions:
\begin{itemize}
\item \textbf{RQ1.} Does conditional generation estimate the hourly PNSD, over all 50 diameter bins, more accurately than deterministic regression, historical means and alternative generative methods?
\item \textbf{RQ2.} Do physical constraints improve the generated PNSDs, and does the form of the constraint matter?
\item \textbf{RQ3.} For two derived quantities with contrasting size weighting, \ncufp{} and CS, are the generated PNSDs more accurate than the PNSDs observed in the same season and at the same hour of day during the training period?
\item \textbf{RQ4.} How much of the skill gained from PM$_{2.5}$ and meteorology carries over when a model trained at one station is applied, without retraining, at the other?
\end{itemize}
Our contributions are a benchmark and a set of diagnostics rather than a single optimized model, and they address the three gaps identified above: single deterministic estimates, the evaluation of derived quantities, and transfer between sites. First, we formulate PNSD estimation as conditional generation, so that the mass--number ambiguity appears as ensemble spread rather than being averaged away, and compare six generative models on equal terms. Second, we compare integral and dynamical physical constraints under one training objective. Third, we evaluate the generated PNSDs through \ncufp{} and CS and under transfer between sites. As the reference, we sample complete PNSD sequences observed in the training period in the same season and at the same hour of day; this historical ensemble shows what could be estimated without any knowledge of the current PM$_{2.5}$ and weather, so improving on it shows what those observations add. Our main findings are: (1) all six generative models have a lower continuous ranked probability score (CRPS) than gradient boosting at both sites; (2) the dynamical constraint lowers the CRPS of the PNSD by 11.5\,\% and 5.5\,\% at the two sites, whereas either integral constraint alone raises it; (3) the GDE-constrained model has the lowest CRPS of the four learned models for both CS and \ncufp{} at both sites; and (4) transferability depends on both the quantity and the direction of transfer, and boreal-to-urban transfer fails for \ncufp. These findings are a proof of concept on a controlled two-station testbed. They show which modelling choices matter and how models trained on a single station fail, which is the information needed before scaling the approach to large, multi-station training data. Section~\ref{sec:data} describes the observations and the task, Sect.~\ref{sec:methods} the model and evaluation, and Sect.~\ref{sec:results} the results, organized by research question. Section~\ref{sec:discussion} interprets the results, lists limitations and outlines future work.

\section{Observations and estimation task}
\label{sec:data}

\subsection{A boreal--urban station pair}
We use SMEAR~II at Hyyti\"al\"a (61.85$^\circ$\,N, 24.29$^\circ$\,E) and SMEAR~III at Kumpula, Helsinki (60.20$^\circ$\,N, 24.96$^\circ$\,E). Hyyti\"al\"a is a boreal forest station where secondary particle formation and growth are prominent features of the particle population \citep{hari2005,dalmaso2005}. Kumpula is an urban background station near a major road in the Helsinki metropolitan area of more than a million inhabitants, with local traffic contributions alongside regional aerosol \citep{jarvi2009}. The two stations therefore represent contrasting situations in which emissions and meteorology shape the observed particle mass and number concentrations. We chose them as a testbed, not as a representative sample of aerosol environments: their seven-year common record allows identical training and evaluation periods for all models, and their contrast provides an out-of-distribution test in each direction.

\subsection{Data}
The data set used in this study covers 1 January 2018--31 December 2024 at both stations, at hourly resolution. PNSDs measured by the DMPS are averaged from native 10\,min measurements. We retain the 50 archived diameter bins between 2.82 and 794\,nm, without interpolation. This grid covers part of the nucleation mode and the Aitken and accumulation modes; all predicted PNSDs and derived quantities are limited to this size range.

The upper limit of 794\,nm has a direct consequence for mass. PM$_{2.5}$ includes particles up to 2.5\,$\mu$m in aerodynamic diameter, which corresponds to a mobility diameter of about 2\,$\mu$m for a density of 1.5\,g\,cm$^{-3}$ (Appendix~\ref{app:physics}). Even the full DMPS range, which ends at 1000\,nm, does not reach this size. The mass integrated from a predicted PNSD is therefore a resolved, partial mass that cannot be required to equal PM$_{2.5}$; Sect.~\ref{sec:physics} explains how we use it only as a soft reference.

The model inputs are PM$_{2.5}$ measured in situ and eight variables of the ERA5 reanalysis at the nearest grid cell \citep{hersbach2020}: 2\,m temperature and dewpoint, 10\,m wind components, surface pressure, boundary-layer height, total precipitation and downward solar radiation. ERA5 is model output constrained by observations rather than a local measurement, but it is available everywhere at hourly resolution. The variables describe the processes that most directly shape the PNSD at a station: boundary-layer height and wind control dilution and transport, solar radiation drives photochemistry and new particle formation, temperature and dewpoint affect condensation and the uptake of water, and precipitation removes particles by wet deposition. In addition, the total number concentration measured by a condensation particle counter (CPC) is used for one training constraint. It agrees closely with the total number integrated from the DMPS PNSD (Sect.~\ref{sec:physics}). PNSDs are log-transformed and normalized per bin (Appendix~\ref{app:data}). 

\subsection{Estimation task}
The task is not a forecast: the model estimates the PNSD for the same hours as its inputs. For a window ending at hour $\tau$, the model receives the PM$_{2.5}$ and ERA5 record of those 24 hours, $\mathbf c_\tau=(\mathrm{PM}_{2.5},\mathrm{ERA5})_{\tau-23:\tau}$, and generates the PNSD for the same 24 hours, $\mathbf x_\tau=\mathrm{PNSD}_{\tau-23:\tau}\in\mathbb R^{50\times24}$ (50 diameter bins by 24 hours). No PNSD measurement is given to the model at any time. A 24\,h window covers one full diurnal cycle of boundary-layer development, radiation and emissions, which strongly shape the PNSD. Each example therefore requires 24 consecutive valid hours. Models are trained separately at each site and evaluated on a later period at that site and, without retraining, at the other site.

\section{Methods}
\label{sec:methods}

\subsection{Conditional generation with flow matching}
Flow Matching (FM) \citep{lipman2023} learns a velocity field $\mathbf v_\theta$ that transports Gaussian noise $\mathbf x_0$ to a PNSD sequence $\mathbf x_1$, conditional on $\mathbf c$. During training, $t$ is sampled uniformly on $[0,1]$, the intermediate state is $\mathbf x_t=(1-t)\mathbf x_0+t\mathbf x_1$, and the target velocity is $\mathbf v^*=\mathbf x_1-\mathbf x_0$:
\begin{equation}
\mathcal{L}_{\mathrm{FM}} =
\mathbb{E}_{t,\mathbf{x}_0,(\mathbf{x}_1,\mathbf c)}
\left\| \mathbf{v}_\theta(\mathbf{x}_t,t,\mathbf{c}) - \mathbf{v}^\ast \right\|^2 .
\label{eq:fm}
\end{equation}
Here $t$ is a generative time that indexes the integration from noise to data; it is distinct from the physical (clock) time $\tau$ of the measurements. FM is not a data-assimilation method such as the Kalman filter: it does not update a state sequentially with new observations, but learns a mapping from noise to PNSDs conditioned on the inputs. At inference, 50 Euler steps integrate the learned field from a fresh Gaussian draw. Different draws give different PNSD sequences for the same input, and eight draws form the ensemble. The network $\mathbf v_\theta$ is a Transformer \citep{vaswani2017} that treats the 24 hourly PNSDs of a window as a sequence: each hour is embedded as one element, and self-attention lets the estimate for every hour draw on the inputs and PNSDs of all other hours. We chose this architecture because the PNSD evolves through the day and is strongly autocorrelated in time, so the PNSD at one hour holds information about the PNSDs at neighbouring hours. The network has approximately $1.24\times10^6$ parameters (Appendix~\ref{app:models}), and Fig.~\ref{fig:trajectory} illustrates the generation path.

\subsection{Physical constraints}
\label{sec:physics}
A generative objective does not by itself make the generated PNSDs obey aerosol physics. We therefore add soft physical penalties to the training objective. Physical relationships are defined for clean PNSDs, not for the partly noised intermediate states $\mathbf x_t$, so we evaluate them on the one-step clean-sample estimate $\hat{\mathbf x}_1=\mathbf x_t+(1-t)\mathbf v_\theta$, obtained by inverting the interpolation, after conversion to physical units. This estimate reuses the network output, so the penalties require no additional network evaluation. The total objective is
\begin{equation}
\mathcal{L} = \mathcal{L}_{\mathrm{FM}}
+ \lambda_{\mathrm{m}}t^2\mathcal{L}_{\mathrm{mass}}
+ \lambda_{\mathrm{n}}t^2\mathcal{L}_{\mathrm{NC}}
+ \lambda_{\mathrm{g}}t^3\mathcal{L}_{\mathrm{GDE}} .
\label{eq:total}
\end{equation}
Near the noise end of the generative path ($t\to0$), $\hat{\mathbf x}_1$ is little more than a guess, and penalizing it would mainly add noise to training. The factors $t^2$ and $t^3$ therefore down-weight the penalties there. The GDE term receives the steeper weight because it involves derivatives in time and size, which amplify any remaining noise. The two integral constraints ($\mathcal{L}_{\mathrm{mass}}$ and $\mathcal{L}_{\mathrm{NC}}$) restrict integrals of the PNSD; the dynamical constraint ($\mathcal{L}_{\mathrm{GDE}}$) links neighbouring sizes and hours. Coefficients and numerical details are given in Appendix~\ref{app:physics}.

\paragraph{Integral constraint 1: pseudo-mass consistency}
Let $\hat n_{h,i}=\mathrm dN/\mathrm d\log_{10}D_{p,i}$ be the estimated number concentration in diameter bin $i$ at hour $h$, normalized by the logarithmic bin width $\Delta_i$, so that $\hat n_{h,i}\Delta_i$ is the number concentration of particles in that bin. Assuming spherical particles of density $\rho=1.5$\,g\,cm$^{-3}$, the mass resolved by the PNSD is
\begin{equation}
\hat M_{\mathrm{res},h}=\frac{\rho\pi}{6}
\sum_{i=1}^{B}D_{p,i}^{3}\hat n_{h,i}\Delta_i,
\qquad B=50 ,
\label{eq:mass}
\end{equation}
which is compared with the observed PM$_{2.5}$, $m_h^{\rm obs}$, through a robust log-space discrepancy:
\begin{equation}
\mathcal L_{\rm mass}=\frac1T\sum_{h=1}^{T}
\operatorname{Huber}_{0.5}\!\left[
\ln(1+\hat M_{\mathrm{res},h}/M_0)
-\ln(1+\max(m_h^{\rm obs},0)/M_0)\right],
\quad M_0=1\,\mu\mathrm{g}\,\mathrm{m}^{-3},
\label{eq:l_mass}
\end{equation}
with $T=24$ hours. For concentrations well above $M_0$, the discrepancy reduces to the logarithm of the mass ratio, $\ln(\hat M_{\mathrm{res},h}/m_h^{\rm obs})$. The penalty is smallest when this ratio is one, although the resolved mass is expected to be smaller than PM$_{2.5}$ by an unknown factor that varies with the share of mass carried by particles between 0.8 and 2.5\,$\mu$m. The Huber function, which is quadratic for small discrepancies and linear for large ones, and a small weight prevent this mismatch from dominating training, so the term acts as a soft reference rather than an equality. We call it \emph{pseudo-mass consistency} and label it ``mass'' in the tables. In this task, the target window coincides with the input window, so $m_h^{\rm obs}$ is the same PM$_{2.5}$ record that the model receives as input. The term is thus a self-consistency penalty between input and output.

\paragraph{Integral constraint 2: total-number consistency}
Mass constrains small particles only weakly, whereas total number is dominated by them. An independent measurement of total number can therefore constrain the part of the PNSD that mass resolves poorly. Let $c_h^{\rm obs}$ be the total number concentration measured by the CPC at hour $h$, and let $\mathcal H_{\rm NC}\subseteq\{1,\ldots,T\}$ be the set of hours in the window with a valid CPC measurement. We compare the total number resolved by the PNSD, $\hat{\mathrm{NC}}_{\mathrm{tot},h}$, with the CPC record:
\begin{equation}
\hat{\mathrm{NC}}_{\mathrm{tot},h}
=\sum_{i=1}^{B}\hat n_{h,i}\Delta_i,
\qquad
\mathcal L_{\rm NC}=\frac{1}{|\mathcal H_{\rm NC}|}
\sum_{h\in\mathcal H_{\rm NC}}
\operatorname{Huber}_{0.5}\!\left[
\ln(1+\hat{\mathrm{NC}}_{\mathrm{tot},h}/N_0)
-\ln(1+\max(c_h^{\rm obs},0)/N_0)\right],
\quad N_0=1\,\mathrm{cm}^{-3} .
\label{eq:l_cn}
\end{equation}
The CPC and DMPS have different detection ranges and responses, so their total number concentrations are closely related but not identical (log-space correlations of 0.988 and 0.963, median DMPS/CPC ratios 0.87 and 1.03 at the two sites; Appendix~\ref{app:physics}). We refer to this constraint as \nctot{} consistency and label it ``number'' in the tables and figures.

\paragraph{Dynamical constraint: reduced general dynamic equation (GDE)}
Aerosol dynamical processes, namely new particle formation, condensational growth, coagulation and deposition, change the particle population over time. Growth and coagulation in particular link neighbouring sizes and successive hours, a structure that the integral constraints cannot express. We also include boundary-layer dilution, which changes the PNSD observed at a fixed station. We penalize the relative residual of a reduced GDE,
\begin{equation}
r_{h,i}=\frac{1}{\hat n_{h,i}+\epsilon}
\left[
\frac{\partial\hat n}{\partial\tau}
+\frac{\partial}{\partial\ln D_p}
\left(\frac{\widehat{\mathrm{GR}}}{D_p}\hat n\right)
-\hat S+\widehat{\mathrm{CoagS}}\,\hat n
+\lambda_{\mathrm{dil}}\hat n
\right]_{h,i},
\label{eq:gde_residual}
\end{equation}
where $\tau$ is physical (clock) time in hours and $\epsilon$ stabilizes the division. The growth rate $\widehat{\mathrm{GR}}$ moves particles through size space, the coagulation sink $\widehat{\mathrm{CoagS}}$ is evaluated on the predicted PNSD, $\hat S$ absorbs unresolved sources and transport, and $\lambda_{\mathrm{dil}}=(\mathrm dH/\mathrm d\tau)_+/H$ is the dilution rate from boundary-layer height $H$. An auxiliary head reads only the conditioning and predicts growth and source terms together with a bounded log residual variance $s_{h,i}$. The variance is needed because a single station cannot close the aerosol budget: on some days, advection of a different air mass dominates every local term, and the model should be able to down-weight such hours rather than be forced to balance them. The penalty is therefore the heteroscedastic Gaussian negative log-likelihood
\begin{equation}
\mathcal L_{\mathrm{GDE}}=\frac{1}{TB}
\sum_{h=1}^{T}\sum_{i=1}^{B}
\left[\exp(-s_{h,i})r_{h,i}^{2}+s_{h,i}\right] ,
\label{eq:l_gde}
\end{equation}
in which the second term prevents the model from lowering the loss merely by inflating the variance. The equation is reduced: new particle formation enters only as a growth flux into the smallest resolved bin and through $\hat S$, deposition and the coagulation gain term are not represented explicitly, and $\hat S$ can absorb these and other omitted processes. It therefore regularizes the generated evolution, but it does not close the aerosol budget at an open measurement site.

\subsection{Comparison models and references}
\label{sec:baselines}
\textit{Generative models.} The five comparison models span the main methods of learning a conditional distribution: iterative denoising, alternative transport couplings, guidance, one-step generation and energy-based modelling. A denoising diffusion model \citep{ho2020} reuses the FM backbone as a noise predictor. FM+OT replaces independent noise--data pairing with minibatch optimal-transport coupling \citep{tong2024}. FM+CFG uses classifier-free guidance \citep{ho2022}, with the tabulated operating point at guidance scale $w=1$. The Drifting Model \citep{deng2026} generates with one network evaluation, and Energy Matching \citep{balcerak2025} learns a scalar potential.

\textit{Deterministic models.} Gradient boosting fits one histogram regressor per diameter bin, and a multilayer perceptron (MLP) predicts all $24\times50$ values jointly. Both return one deterministic PNSD per input, and so show what is gained or lost when the ambiguity is not represented.

\textit{Non-learning references.} For the PNSD, the reference is the mean PNSD of the training period for each calendar month and hour of day. We label it ``climatology'' in the tables and figures, but note that it averages seven years rather than the 30 years of a meteorological climatology. For \ncufp{} and CS, we use a deterministic \emph{seasonal template} (the mean PNSD by season and UTC hour) and a probabilistic \emph{historical ensemble}. Each member of the historical ensemble is a complete 24\,h PNSD sequence from the training period, matched to the target season and starting hour, so that it retains realistic dependence across bins and hours. It uses no PM$_{2.5}$ or meteorology and therefore measures directly what the conditioning adds. All references are built from source-site training data only.

\subsection{Evaluation design and fairness}
\label{sec:eval}\label{sec:splits}
Training ends at 2 November 2024, 00:00\,UTC. The temporal holdout covers 2 November--31 December 2024 and contains 1417 hourly-stride windows per site. Table~\ref{tab:protocol} lists the three evaluation populations. In the paired transfer test, a model trained at the other station, applied without retraining, and a locally trained model are scored on identical target-site windows, so their ratio isolates the cost of transfer. For a fair comparison, all learned models share the same inputs, target grid, training period and evaluation windows. Since the purpose is a controlled comparison of model families and constraints, we keep these settings fixed rather than tuning each model to its best achievable score on the test data; the reported scores therefore indicate relative performance under identical conditions, not the upper limit of any method.

\begin{table*}[!htbp]
\caption{Evaluation populations. H and K denote Hyyti\"al\"a and Kumpula. All predictions cover the same 24 hours as their input record.}
\label{tab:protocol}
\centering\small
\begin{tabular}{p{0.23\textwidth}p{0.34\textwidth}p{0.33\textwidth}}
\tophline
Test & Scoring population & Question \\
\middlehline
Same-site temporal holdout & 1417 windows at each site; 2 November--31 December 2024 & Local skill on a later period (RQ1--RQ3) \\
Paired transfer & H-trained and K-trained models on the identical target-site holdout & Cost of transfer for \ncufp{} and CS (RQ4) \\
Extended PNSD transfer & All valid windows at the other site & Robustness over a longer period (Appendix~\ref{app:transfer}) \\
\bottomhline
\end{tabular}
\end{table*}

\subsection{Metrics}
Table~\ref{tab:metrics} summarizes the metrics. Each ensemble member is converted to physical units before scoring. For evaluated bin--hour pairs $j=(w,h,i)$, indexed by window $w$, the ensemble mean over $M=8$ members is $\bar n_j$, and the root mean square error over $Q$ pairs is
\begin{equation}
\bar n_j=\frac1M\sum_{m=1}^{M}\hat n_j^{(m)},
\qquad
\mathrm{RMSE}=
\left[\frac1Q\sum_{j}(\bar n_j-n_j^{\rm obs})^2\right]^{1/2} .
\label{eq:rmse}
\end{equation}
The CRPS \citep{gneiting2007} scores the whole ensemble; we use its finite-ensemble (fair) form
\begin{equation}
\mathrm{CRPS}_j=\frac{1}{M}\sum_{m=1}^{M}|z_m-y|
-\frac{1}{2M(M-1)}\sum_{m=1}^{M}\sum_{k\neq m}|z_m-z_k|,
\label{eq:crps}
\end{equation}
with $y=n_j^{\rm obs}$ and $z_m=\hat n_j^{(m)}$, averaged over all pairs. The first term penalizes distance from the observation; the second credits spread, so that spread alone cannot improve the score. For a deterministic prediction, CRPS equals the absolute error. We refer to scores computed over all 50 bins and 24 hours as PNSD scores. The skill score relative to the historical ensemble is
\begin{equation}
\mathrm{CRPSS}=1-
\overline{\mathrm{CRPS}}_{\rm model}/
\overline{\mathrm{CRPS}}_{\rm historical} .
\label{eq:crpss}
\end{equation}
Interval reliability uses the empirical 10th and 90th percentiles $q_{0.1,j}$ and $q_{0.9,j}$:
\begin{equation}
C_{80}=\frac1Q\sum_{j}
\mathbf1\{q_{0.1,j}\leq y_j\leq q_{0.9,j}\},
\qquad
W_{80}=\frac1Q\sum_{j}(q_{0.9,j}-q_{0.1,j}) .
\label{eq:coverage}
\end{equation}
The indicator function $\mathbf1\{\cdot\}$ counts an observation when it lies inside the interval. A value $C_{80}=0.60$, for example, means that only 60\,\% of observations fall inside a nominal 80\,\% interval. Additional results are in Appendix~\ref{app:metrics}.

\begin{table*}[!htbp]
\caption{Metrics, their interpretation and optimal values. RMSE and CRPS carry the units of the scored quantity.}
\label{tab:metrics}
\centering\small
\begin{tabular}{p{0.17\textwidth}p{0.5\textwidth}p{0.25\textwidth}}
\tophline
Metric & What it measures & Optimum \\
\middlehline
RMSE, Eq.~(\ref{eq:rmse}) & Point error of the ensemble mean, weighting large concentration errors & 0 (lower is better) \\
corr. & Pearson correlation of log-transformed PNSDs across bins and hours within a window, averaged over windows & 1 \\
CRPS, Eq.~(\ref{eq:crps}) & Accuracy of the whole predictive distribution & 0 (lower is better) \\
CRPSS, Eq.~(\ref{eq:crpss}) & Improvement over the historical ensemble & 1; 0 equals the reference \\
$C_{80}$, Eq.~(\ref{eq:coverage}) & Fraction of observations inside the central 80\,\% interval & 0.80 \\
$W_{80}$, Eq.~(\ref{eq:coverage}) & Mean width of that interval; read with $C_{80}$ & Narrowest width at $C_{80}=0.80$ \\
Brier score & Mean squared error of an event probability (Appendix~\ref{sec:res_episode}) & 0 \\
\bottomhline
\end{tabular}
\end{table*}

\paragraph{Derived atmospheric quantities}
As motivated in the Introduction, we evaluate two quantities with contrasting diameter weighting. The first is \ncufp, the number concentration of particles over the observed range 2.82--100\,nm. Its upper bound follows the conventional definition of ultrafine particles as those smaller than 100\,nm \citep{who2021}; its lower bound is set by the DMPS grid. The second is the resolved CS \citep{kulmala2001},
\begin{equation}
\mathrm{CS} = 2\pi D \sum_i \beta_m(D_{p,i})\,D_{p,i}\,N_i ,
\label{eq:cs}
\end{equation}
where $N_i=n_i\Delta\log_{10}D_{p,i}$, $D$ is the vapor diffusion coefficient and $\beta_m$ the Fuchs--Sutugin correction. We use sulfuric-acid vapor properties, unit accommodation, 293.15\,K and 101325\,Pa for both predicted and observed PNSDs, so that errors reflect the PNSD only. \ncufp{} counts every particle equally, whereas CS weights each particle by its diameter and a transition-regime correction, so that larger particles contribute more. Each ensemble member is integrated separately before scoring, which preserves the dependence across sizes.

\begin{figure*}[t]
\includegraphics[width=\textwidth]{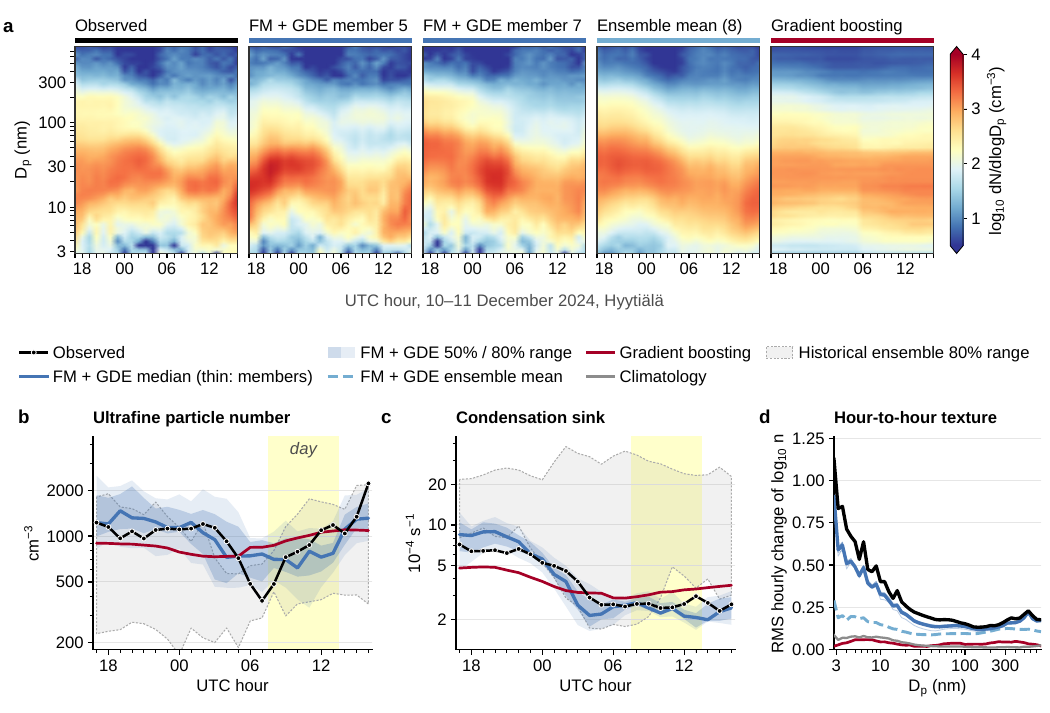}
\caption{\textbf{Individual ensemble members retain the hour-to-hour structure of the PNSD that the ensemble mean, gradient boosting and the climatology smooth out.} (a)~PNSDs for the temporal-holdout window from 10 December 2024, 17:00\,UTC, to 11 December 2024, 16:00\,UTC, at Hyyti\"al\"a: observation, two of the eight FM+GDE members (the two closest to the ensemble median of window-mean \ncufp), the mean of the eight members and gradient boosting. The window was selected automatically, not by hand. Color is $\log_{10}\,\mathrm{d}N/\mathrm{d}\log D_p$. (b)~\ncufp{} and (c)~CS in the same window: observation (black dots), median of the eight FM+GDE members (blue line) with their central 50\,\% and 80\,\% ranges (shading), gradient boosting (red) and the central 80\,\% range of the historical ensemble (gray, dashed outline); the yellow band marks daylight hours (ERA5 downward solar radiation above zero). (d)~Hour-to-hour variability over 60 non-overlapping holdout windows: root-mean-square change of $\log_{10}\mathrm dN/\mathrm d\log D_p$ between consecutive hours, per diameter bin, for the observation (black), the individual FM+GDE members (blue; thin lines, each member), their mean (dashed), gradient boosting and the climatology.}
\label{fig:realism}
\end{figure*}

\section{Results}
\label{sec:results}
We first inspect the generated PNSDs and their scores (RQ1), then the effect of physical constraints (RQ2), \ncufp{} and CS (RQ3) and transfer (RQ4). Finally, we place the generated PNSDs in the context of a process model. Throughout, the aim is to compare model families and constraints under identical conditions and to identify where they fail.

\subsection{RQ1: Generative ensembles give the best distributional skill, while gradient boosting remains competitive on point error}
\label{sec:res_interpolation}\label{sec:res_calibration}

Before scoring, we check whether the generated PNSDs look like observed ones (Fig.~\ref{fig:realism}). In the example window at Hyyti\"al\"a, individual FM+GDE members reproduce the Aitken mode and its weakening in the early morning, as well as the growth of small particles in the afternoon. The ensemble mean is smoother, and gradient boosting returns bands that barely change through the day (Fig.~\ref{fig:realism}a). The ensemble median follows the observed \ncufp{} and CS through most of the window, and its central 80\,\% range is much narrower than that of the historical ensemble (Fig.~\ref{fig:realism}b,c). Over 60 non-overlapping holdout windows, individual members retain most of the observed hour-to-hour variability of the PNSD, whereas the ensemble mean, gradient boosting and the climatology retain only a small part of it, especially for the smallest particles (Fig.~\ref{fig:realism}d). The ensemble mean is therefore not itself a realistic PNSD, and scores that use only the mean understate what the ensemble represents. Over the full holdout, FM retains 83\,\% of the observed standard deviation of the PNSD between windows, whereas gradient boosting retains 58\,\% (Appendix~\ref{app:diagnostics}); one generated PNSD from every model is shown for one holdout day in Fig.~\ref{fig:grid_extrap}.

\begin{figure*}[t]
\centering
\includegraphics[width=\textwidth]{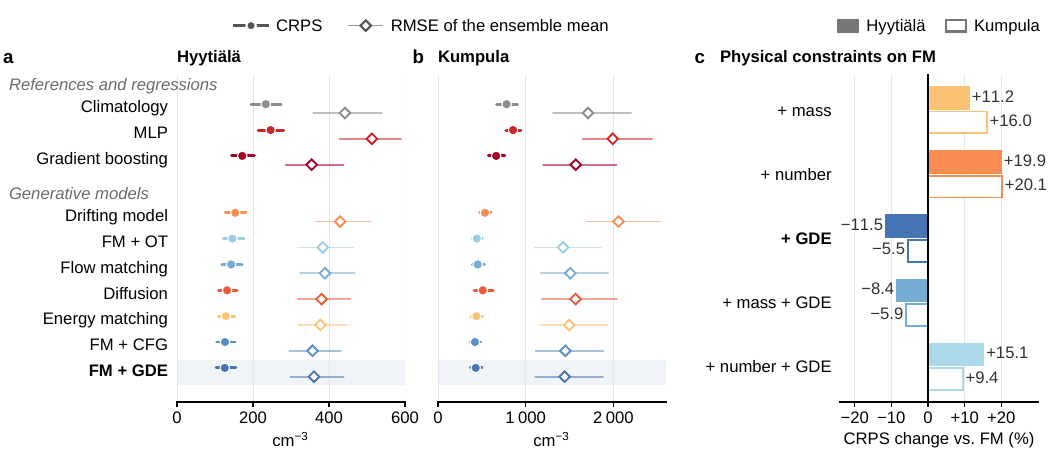}
\caption{\textbf{Generative ensembles have lower CRPS than gradient boosting at both sites, and the GDE constraint is the only physical constraint that improves on unconstrained FM.} (a)~Hyyti\"al\"a and (b)~Kumpula: CRPS (filled circles) and RMSE of the ensemble mean (open diamonds) of the PNSD, both in cm$^{-3}$ of $\mathrm dN/\mathrm d\log_{10}D_p$, with 95\,\% block-bootstrap intervals on the temporal holdouts (1417 windows per site). Deterministic models provide a single prediction, and FM+GDE (highlighted) comes from the constraint ablation of Sect.~\ref{sec:res_physics}. (c)~Change in CRPS relative to unconstrained FM for each physical constraint (filled, Hyyti\"al\"a; open, Kumpula); negative values are improvements, and ``number'' denotes the \nctot{} constraint.}
\label{fig:skill}
\end{figure*}

Figure~\ref{fig:skill}a,b and Table~\ref{tab:performance} give the PNSD scores. All six generative models have lower CRPS than gradient boosting and the climatology at both sites. FM+CFG has the lowest CRPS among the unconstrained models, 126\,cm$^{-3}$ at Hyyti\"al\"a and 421\,cm$^{-3}$ at Kumpula, compared with 142\,cm$^{-3}$ and 453\,cm$^{-3}$ for plain FM.

Point scores tell a more mixed story. At Hyyti\"al\"a, gradient boosting has the lowest RMSE (354\,cm$^{-3}$, against 389\,cm$^{-3}$ for FM) and the highest correlation (0.862, against 0.820). This advantage is expected: gradient boosting minimizes the squared error in each bin and therefore returns the conditional mean, which is exactly what RMSE rewards. At Kumpula, FM+OT has the lowest RMSE (1427\,cm$^{-3}$), and five of the six generative models are below gradient boosting (1571\,cm$^{-3}$).

For the constraint and transfer experiments that follow, we retain plain FM as the base model. Guidance and optimal-transport coupling change the sampling and the training pairs, respectively, independently of the physical penalties, so they could be combined with the constraints later; using the plain model keeps the effect of each constraint isolated. The guidance trade-off is analysed in Appendix~\ref{sec:res_cfg}.

\begin{table*}[!htbp]
\caption{\textbf{PNSD scores on the same-site temporal holdouts.} corr.\ uses log-transformed PNSDs; CRPS and RMSE are in cm$^{-3}$, with 95\,\% block-bootstrap intervals for CRPS. ``Climatology'' is the month--hour mean of the training period. Best values are in bold.}
\label{tab:performance}
\centering
\begin{minipage}[t]{0.49\textwidth}
\centering
\textbf{(a) Hyyti\"al\"a}\par
\scriptsize
\begin{tabular}{lrrr}
\tophline
Model & corr. $\uparrow$ & CRPS $\downarrow$ & RMSE $\downarrow$ \\
\middlehline
Climatology & 0.776 & 234 (195--274) & 442 \\
Gradient Boosting & \textbf{0.862} & 172 (144--204) & \textbf{354} \\
MLP & 0.783 & 246 (213--281) & 513 \\
Diffusion Model & 0.771 & 132 (110--158) & 382 \\
Drifting Model & 0.779 & 152 (127--180) & 425 \\
Energy Matching & 0.821 & 129 (110--152) & 376 \\
Flow Matching & 0.820 & 142 (118--171) & 389 \\
FM + OT & 0.822 & 146 (122--175) & 383 \\
FM + CFG & 0.826 & \textbf{126 (104--153)} & 357 \\
\bottomhline
\end{tabular}
\end{minipage}\hfill
\begin{minipage}[t]{0.49\textwidth}
\centering
\textbf{(b) Kumpula}\par
\scriptsize
\begin{tabular}{lrrr}
\tophline
Model & corr. $\uparrow$ & CRPS $\downarrow$ & RMSE $\downarrow$ \\
\middlehline
Climatology & 0.889 & 782 (667--906) & 1710 \\
Gradient Boosting & 0.915 & 663 (575--762) & 1571 \\
MLP & 0.887 & 856 (774--946) & 1993 \\
Diffusion Model & 0.926 & 511 (412--628) & 1568 \\
Drifting Model & 0.888 & 535 (475--606) & 2059 \\
Energy Matching & 0.925 & 437 (375--508) & 1496 \\
Flow Matching & 0.929 & 453 (391--529) & 1509 \\
FM + OT & \textbf{0.932} & 442 (389--505) & \textbf{1427} \\
FM + CFG & \textbf{0.932} & \textbf{421 (366--488)} & 1452 \\
\bottomhline
\end{tabular}
\end{minipage}
\end{table*}

\subsection{RQ2: A dynamical constraint improves the generated PNSDs; integral constraints alone do not}
\label{sec:res_physics}
Figure~\ref{fig:skill}c and Table~\ref{tab:physics_indomain} show that the form of the constraint matters more than its presence. The GDE constraint lowers CRPS from 141.9\,cm$^{-3}$ to 125.7\,cm$^{-3}$ ($-11.5$\,\%) at Hyyti\"al\"a and from 455.3\,cm$^{-3}$ to 430.5\,cm$^{-3}$ ($-5.5$\,\%) at Kumpula, and it also lowers the RMSE of the PNSD and raises the correlation at both sites. The integral constraints have the opposite effect when used alone: pseudo-mass consistency raises CRPS to 157.8\,cm$^{-3}$ ($+11$\,\%) at Hyyti\"al\"a and 528.3\,cm$^{-3}$ ($+16$\,\%) at Kumpula, and \nctot{} consistency raises it by 20\,\% at both sites. Combining them with GDE recovers part of this loss. Mass+GDE gives the lowest CRPS and RMSE at Kumpula (428.4\,cm$^{-3}$ and 1430.0\,cm$^{-3}$, respectively), whereas GDE alone remains best at Hyyti\"al\"a, and the \nctot{}+GDE combination does not reach the unconstrained CRPS at either site.

Within these formulations, therefore, constraining an integral is not enough to recover how the corresponding quantity is distributed across size and time. A mismatch in measurement support also explains why the integral constraints need care. The predicted PNSD ends at 794\,nm, whereas the PM$_{2.5}$ instrument also samples larger particles, so forcing agreement between the two can push the PNSD in the wrong direction. Moreover, the share of PM$_{2.5}$ carried by particles smaller than 0.8\,$\mu$m varies with sources and season, so there is no fixed target ratio towards which the resolved mass could be tuned. At Hyyti\"al\"a, pseudo-mass consistency alone even raises the mass RMSE from 0.19\,$\mu$g\,m$^{-3}$ to 0.22\,$\mu$g\,m$^{-3}$, while GDE alone lowers it to 0.18\,$\mu$g\,m$^{-3}$. Because the mass reference is the input PM$_{2.5}$ itself (Sect.~\ref{sec:physics}), these mass-RMSE values measure input--output consistency rather than independent accuracy. Total number is better matched: CPC number correlates closely with the total number integrated from the DMPS PNSD, although the median DMPS/CPC ratios of 0.87 and 1.03 show that the two instruments still differ (Sect.~\ref{sec:physics}).

\begin{table*}[!htbp]
\caption{\textbf{Physical-constraint ablation on Flow Matching for the same-site temporal holdouts.} CRPS and PNSD RMSE are in cm$^{-3}$; mass RMSE is in $\mu$g\,m$^{-3}$ and measures consistency with the input PM$_{2.5}$. ``Number'' denotes the \nctot{} constraint. Best values are in bold.}
\label{tab:physics_indomain}
\centering
\begin{minipage}[t]{0.49\textwidth}
\centering
\textbf{(a) Hyyti\"al\"a}\par
\scriptsize
\begin{tabular}{lrrrr}
\tophline
Model & corr. & CRPS & PNSD RMSE & mass RMSE \\
\middlehline
Flow Matching & 0.821 & 141.9 & 388.5 & 0.19 \\
FM + mass & 0.808 & 157.8 & 399.6 & 0.22 \\
FM + GDE & \textbf{0.834} & \textbf{125.7} & \textbf{360.5} & 0.18 \\
FM + mass + GDE & 0.828 & 130.0 & 367.3 & 0.18 \\
FM + number & 0.820 & 170.2 & 433.0 & 0.20 \\
FM + number + GDE & 0.832 & 163.4 & 455.0 & \textbf{0.16} \\
\bottomhline
\end{tabular}
\end{minipage}\hfill
\begin{minipage}[t]{0.49\textwidth}
\centering
\textbf{(b) Kumpula}\par
\scriptsize
\begin{tabular}{lrrrr}
\tophline
Model & corr. & CRPS & PNSD RMSE & mass RMSE \\
\middlehline
Flow Matching & 0.929 & 455.3 & 1515.9 & 1.18 \\
FM + mass & 0.917 & 528.3 & 1536.1 & 1.92 \\
FM + GDE & \textbf{0.934} & 430.5 & 1444.4 & 1.14 \\
FM + mass + GDE & 0.933 & \textbf{428.4} & \textbf{1430.0} & \textbf{0.77} \\
FM + number & 0.918 & 547.1 & 1577.0 & 1.28 \\
FM + number + GDE & 0.930 & 498.3 & 1567.8 & 1.36 \\
\bottomhline
\end{tabular}
\end{minipage}
\end{table*}

\subsection{RQ3: Routine observations add information beyond historical PNSDs}
\label{sec:res_atmquant}
\begin{figure*}[t]
\centering
\includegraphics[width=\textwidth]{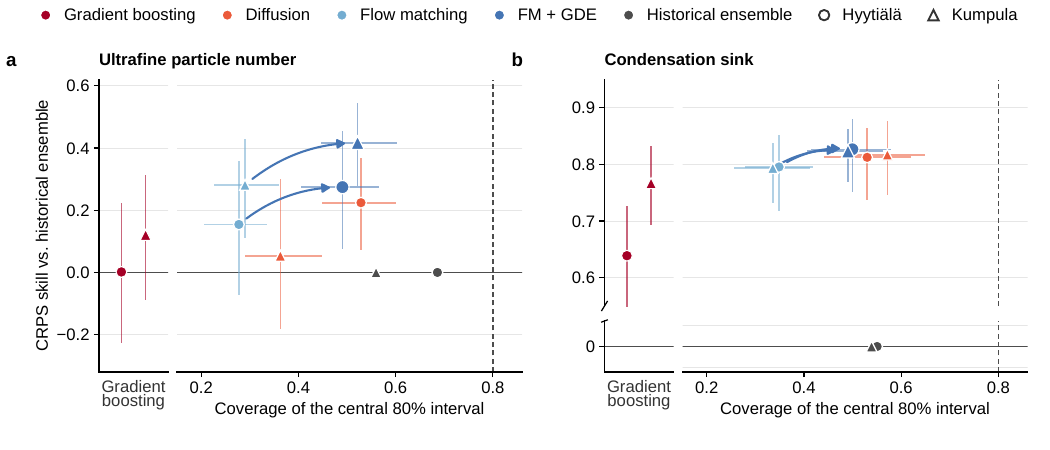}
\caption{\textbf{Locally trained models add skill beyond the historical ensemble for \ncufp{} and CS, but their intervals remain too narrow.} Skill and reliability with local training for (a)~\ncufp{} (2.82--100\,nm) and (b)~the resolved CS. Vertical axis: CRPS skill score relative to the historical ensemble (zero; one is perfect). Horizontal axis: fraction of observations inside the central 80\,\% ensemble interval (nominal 0.80, dashed). Circles, Hyyti\"al\"a; triangles, Kumpula; crosses, 95\,\% paired block-bootstrap intervals over the 1417 holdout windows. Arrows join plain FM to FM+GDE at each site. Gradient boosting has no interval and is shown in its own column; panel b uses a broken axis.}
\label{fig:quantities}
\end{figure*}

A model can improve a bin-wise score without improving a derived quantity. We therefore compare \ncufp{} and CS with the historical ensemble, which already contains realistic PNSDs and their seasonal variation (Fig.~\ref{fig:quantities}, Table~\ref{tab:paired_quantity_summary}). Every learned model matches or improves on this reference at both sites (CRPSS $\geq0$; Table~\ref{tab:atmquant}). The weakest case is gradient boosting at Hyyti\"al\"a, which only matches the historical ensemble for \ncufp{} (CRPSS 0.00) without improving on it; at Kumpula, gradient boosting estimates \ncufp{} better than diffusion. FM+GDE has the lowest CRPS of the four learned models in all four site--quantity comparisons. Its CRPSS is 0.27 and 0.42 for \ncufp{} and 0.83 and 0.82 for CS at Hyyti\"al\"a and Kumpula, respectively.

Most of the improvement comes from conditioning itself. For \ncufp, the CRPS falls from 250\,cm$^{-3}$ for the historical ensemble to 211\,cm$^{-3}$ for FM at Hyyti\"al\"a, and from 1028\,cm$^{-3}$ to 739\,cm$^{-3}$ at Kumpula. GDE then lowers it further, to 181\,cm$^{-3}$ ($-14$\,\%) and 599\,cm$^{-3}$ ($-19$\,\%). The PM$_{2.5}$ and meteorological record therefore carries information on the PNSD beyond its seasonal and diurnal pattern.

The intervals are informative but too narrow. For \ncufp, FM+GDE's mean 80\,\% interval is 391\,cm$^{-3}$ wide at Hyyti\"al\"a and 1583\,cm$^{-3}$ at Kumpula, against 1158\,cm$^{-3}$ and 4070\,cm$^{-3}$ for the historical ensemble, but it contains only 49\,\% and 52\,\% of the observations; for CS, coverage is 50\,\% and 49\,\%. Notably, the GDE constraint widens plain FM's intervals and raises its coverage for \ncufp{} from 28\,\% to 49\,\% at Hyyti\"al\"a and from 29\,\% to 52\,\% at Kumpula, shown as arrows in Fig.~\ref{fig:quantities}. The constraint thus improves calibration as well as accuracy, although no learned model reaches the nominal 80\,\%.

\begin{table*}[t]
\caption{\textbf{\ncufp{} and CS scores for local and transferred models on identical target-site windows.} CRPS and central-80\,\% coverage ($C_{80}$, nominal 0.80) are given for each quantity. Interval widths and CRPSS are in Table~\ref{tab:atmquant}.}
\label{tab:paired_quantity_summary}
\centering
\begin{minipage}[t]{0.49\textwidth}
\centering
\textbf{(a) Hyyti\"al\"a target}\par
\renewcommand{\baselinestretch}{1}\small
\setlength{\tabcolsep}{3pt}
\begin{tabular*}{\linewidth}{@{\extracolsep{\fill}}lrrrr@{}}
\tophline
& \multicolumn{2}{c}{\ncufp} & \multicolumn{2}{c}{CS} \\
Model & CRPS & $C_{80}$ & CRPS & $C_{80}$ \\
& (cm$^{-3}$) & & ($10^{-4}$\,s$^{-1}$) & \\
\middlehline
\multicolumn{5}{l}{\textit{Local training}} \\
Gradient boosting & 250 & -- & 1.73 & -- \\
Diffusion & 194 & 0.53 & 0.90 & 0.53 \\
FM & 211 & 0.28 & 0.98 & 0.35 \\
FM + GDE & 181 & 0.49 & 0.83 & 0.50 \\
Historical ensemble & 250 & 0.69 & 4.80 & 0.55 \\
\middlehline
\multicolumn{5}{l}{\textit{Kumpula-trained $\to$ Hyyti\"al\"a}} \\
Gradient boosting & 1378 & -- & 2.08 & -- \\
Diffusion & 532 & 0.21 & 1.33 & 0.28 \\
FM & 1281 & 0.05 & 2.49 & 0.11 \\
FM + GDE & 891 & 0.11 & 1.97 & 0.20 \\
Historical ensemble & 2010 & 0.05 & 11.33 & 0.23 \\
\bottomhline
\end{tabular*}
\end{minipage}\hfill
\begin{minipage}[t]{0.49\textwidth}
\centering
\textbf{(b) Kumpula target}\par
\renewcommand{\baselinestretch}{1}\small
\setlength{\tabcolsep}{3pt}
\begin{tabular*}{\linewidth}{@{\extracolsep{\fill}}lrrrr@{}}
\tophline
& \multicolumn{2}{c}{\ncufp} & \multicolumn{2}{c}{CS} \\
Model & CRPS & $C_{80}$ & CRPS & $C_{80}$ \\
& (cm$^{-3}$) & & ($10^{-4}$\,s$^{-1}$) & \\
\middlehline
\multicolumn{5}{l}{\textit{Local training}} \\
Gradient boosting & 904 & -- & 1.97 & -- \\
Diffusion & 973 & 0.36 & 1.54 & 0.57 \\
FM & 739 & 0.29 & 1.74 & 0.34 \\
FM + GDE & 599 & 0.52 & 1.49 & 0.49 \\
Historical ensemble & 1028 & 0.56 & 8.44 & 0.54 \\
\middlehline
\multicolumn{5}{l}{\textit{Hyyti\"al\"a-trained $\to$ Kumpula}} \\
Gradient boosting & 2294 & -- & 6.26 & -- \\
Diffusion & 1914 & 0.07 & 4.20 & 0.16 \\
FM & 1896 & 0.07 & 3.49 & 0.19 \\
FM + GDE & 1815 & 0.10 & 3.26 & 0.28 \\
Historical ensemble & 1693 & 0.22 & 5.33 & 0.73 \\
\bottomhline
\end{tabular*}
\end{minipage}
\end{table*}

\subsection{RQ4: Transferability depends on both the quantity and the direction of transfer}
\label{sec:res_ood}
\begin{figure*}[!htbp]
\centering
\includegraphics[width=\textwidth]{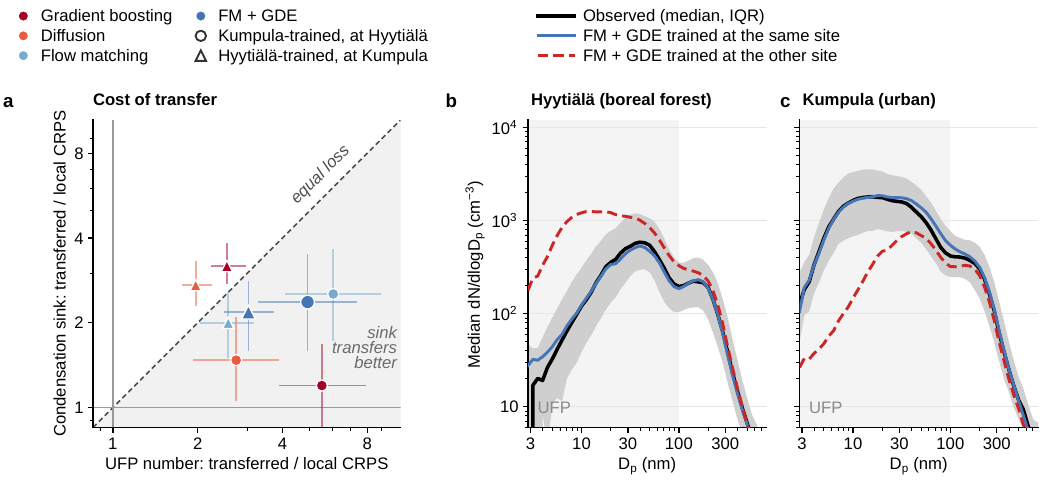}
\caption{\textbf{Transfer between the boreal and the urban site costs more skill for \ncufp{} than for CS, and the error lies in the ultrafine range of the PNSD.} (a)~Cost of transfer: CRPS of a model applied without retraining at the other site, divided by the CRPS of the same model trained locally, on identical target-site windows, for \ncufp{} (horizontal; ``UFP number'') and CS (vertical); circles, Kumpula-trained models evaluated at Hyyti\"al\"a; triangles, Hyyti\"al\"a-trained models evaluated at Kumpula; crosses, 95\,\% paired block-bootstrap intervals. One means no loss, and points below the diagonal lose less skill for CS than for \ncufp. (b,c)~Median PNSD over 60 non-overlapping holdout windows at (b)~Hyyti\"al\"a and (c)~Kumpula: observation (black; gray band, interquartile range) and the eight-member FM+GDE ensembles trained at the same site (blue) or at the other site (red, dashed); the shaded diameter range marks ultrafine particles.}
\label{fig:transfer_cost}
\end{figure*}

A station without its own PNSD measurements would have to rely on a model trained elsewhere. Transfer between our two stations is the most demanding test in this benchmark, because each model has seen only one aerosol regime during training. We use it as a stress test that reveals how single-station training breaks down, not as an estimate of what a model trained on many stations could achieve. To measure the cost of transfer, we apply each model, without retraining, to the other site and divide its CRPS by that of the corresponding locally trained model on the same windows (Fig.~\ref{fig:transfer_cost}a). A ratio of one means that transfer costs nothing.

\textit{Urban to boreal.} From Kumpula to Hyyti\"al\"a, CS degrades less than \ncufp{} for all four learned models, with CRPS ratios of 1.2--2.5 against 2.7--6.1, so all four points lie below the diagonal in Fig.~\ref{fig:transfer_cost}a. Diffusion transfers best: its CRPS for CS rises 1.48-fold, from $0.90\times10^{-4}$\,s$^{-1}$ to $1.33\times10^{-4}$\,s$^{-1}$, while its CRPS for \ncufp{} rises 2.74-fold, from 194\,cm$^{-3}$ to 532\,cm$^{-3}$. Plain FM transfers worse on both counts, with a 6.07-fold degradation for \ncufp{} and a 2.54-fold degradation for CS. Good point accuracy after transfer does not imply good calibration: diffusion's transferred coverage for CS falls to 0.28, from 0.53 locally.

\textit{Boreal to urban.} From Hyyti\"al\"a to Kumpula, CS and \ncufp{} degrade by comparable amounts, with no consistent ordering between them, and this direction shows the more severe failure. For \ncufp, every Hyyti\"al\"a-trained model is less skilful than the Hyyti\"al\"a-derived historical ensemble itself (CRPSS $-0.07$ to $-0.36$; Table~\ref{tab:atmquant}c), and coverage collapses to 0.07--0.10. Plain FM's CRPS for \ncufp{} rises from 739\,cm$^{-3}$ with local training to 1896\,cm$^{-3}$ after transfer. In the example of Fig.~\ref{fig:grid_ood}, no transferred model reproduces the observed daytime UFP enhancement at Kumpula, most likely caused by traffic emissions. CS fares better: transferred FM+GDE has a CRPSS of $+0.39$, although its CRPS ($3.26\times10^{-4}$\,s$^{-1}$) remains more than twice that of local FM+GDE ($1.49\times10^{-4}$\,s$^{-1}$).

\textit{Where in the PNSD transfer fails.} The median PNSDs show that the transfer error lies almost entirely in the ultrafine range (Fig.~\ref{fig:transfer_cost}b,c). At Hyyti\"al\"a, Kumpula-trained FM+GDE overestimates the median PNSD below about 100\,nm, by up to an order of magnitude near 10\,nm. At Kumpula, Hyyti\"al\"a-trained FM+GDE underestimates it in the same range by a similar factor. Above about 200\,nm, transferred and locally trained ensembles agree closely with the observations at both sites. Each transferred model thus carries the ultrafine population of its training station to the other station, while the accumulation mode, which dominates CS, is reproduced in both directions.

\textit{Effect of the constraint.} GDE lowers FM's transferred CRPS in all four direction--quantity cases, by 4--30\,\%, but it does not close the gap to local training in any of them. Skill scores against a transferred reference must also be read with care. In the transfer test, the historical ensemble is built from the training data of the source station, so in the urban-to-boreal direction it consists of Kumpula PNSDs evaluated at Hyyti\"al\"a. These urban PNSDs describe Hyyti\"al\"a poorly: their CRPS for \ncufp{} at Hyyti\"al\"a is 2010\,cm$^{-3}$, against 250\,cm$^{-3}$ for the historical ensemble built from Hyyti\"al\"a's own training data. A transferred model can therefore reach a large positive CRPSS simply because this reference is poor, not because transfer succeeded. The ratio against local training, which compares two models on the same windows, avoids this ambiguity. Scores over all valid windows, fitted-component ratios and complete diagnostics for \ncufp{} and CS are given in Appendices~\ref{app:transfer}--\ref{sec:modal}.

\subsection{Comparison with the SOSAA-FP process model}
\label{sec:sosaa}\label{sec:res_sosaa}
To place the generated PNSDs in context, we compare them with SOSAA-FP, a Lagrangian model that explicitly represents chemistry and aerosol microphysics \citep{boy2011,clusius2026}. We use the SOSAA-FP runs of \citet{clusius2026} for 1 March 2018, 02:00\,UTC to 1 November 2018, 01:00\,UTC (5880 hourly rows), whose size grid covers our 50 diameters (Appendix~\ref{app:sosaa_seasonal}). The matched windows are drawn from the stratified interpolation holdout, which shares observations with neighbouring training windows.

\begin{figure*}[!htb]
\includegraphics[width=0.9\textwidth]{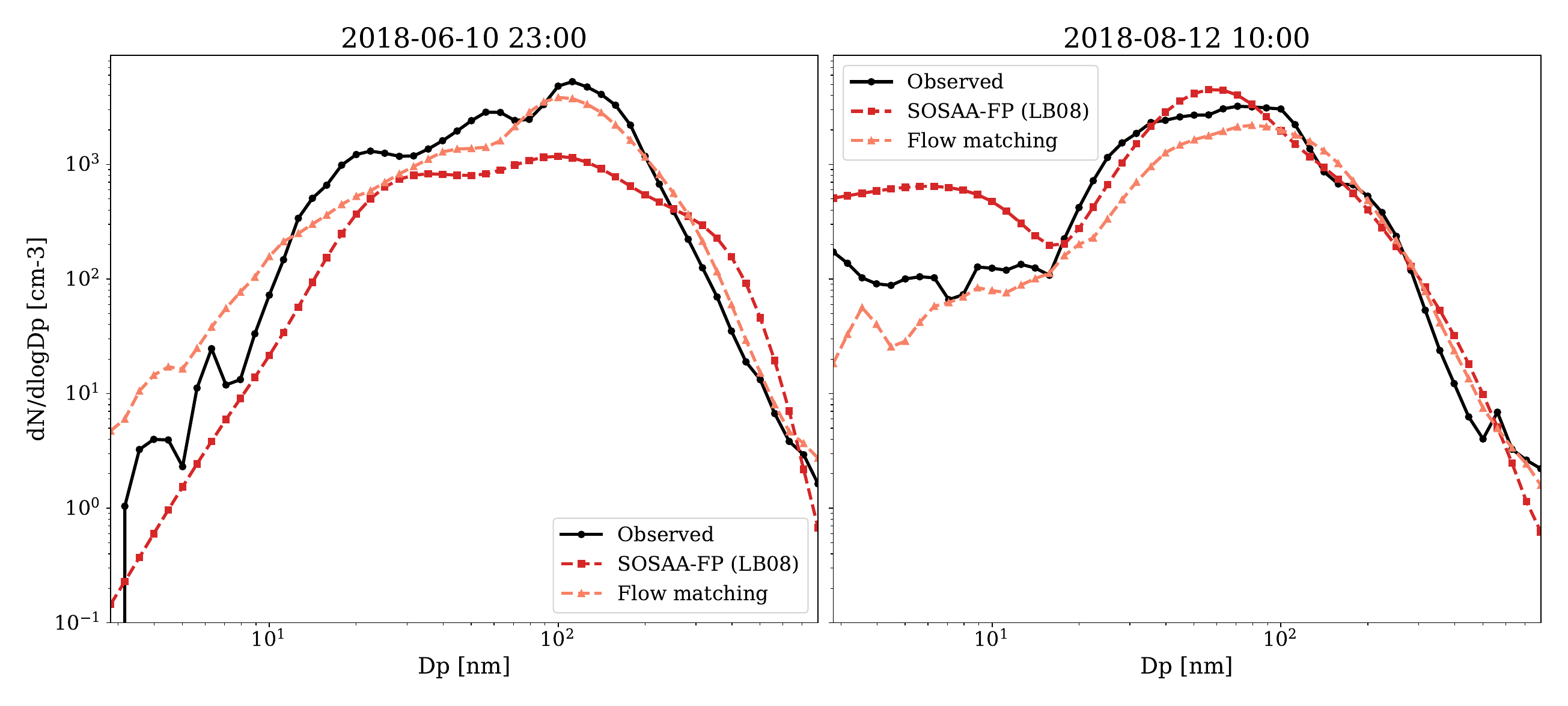}
\caption{\textbf{In two matched cases, Flow Matching follows the observed PNSD amplitude, whereas SOSAA-FP is about an order of magnitude lower.} Hyyti\"al\"a, (a)~10~June 2018, 23:00\,UTC, and (b)~12~August 2018, 10:00\,UTC. Black: observed DMPS PNSD; salmon dashed: Flow Matching; red dashed: SOSAA-FP. Both use the common 2.82--794\,nm interval.}
\label{fig:sosaa_examples}
\end{figure*}

\begin{figure*}[!htb]
\includegraphics[width=0.9\textwidth]{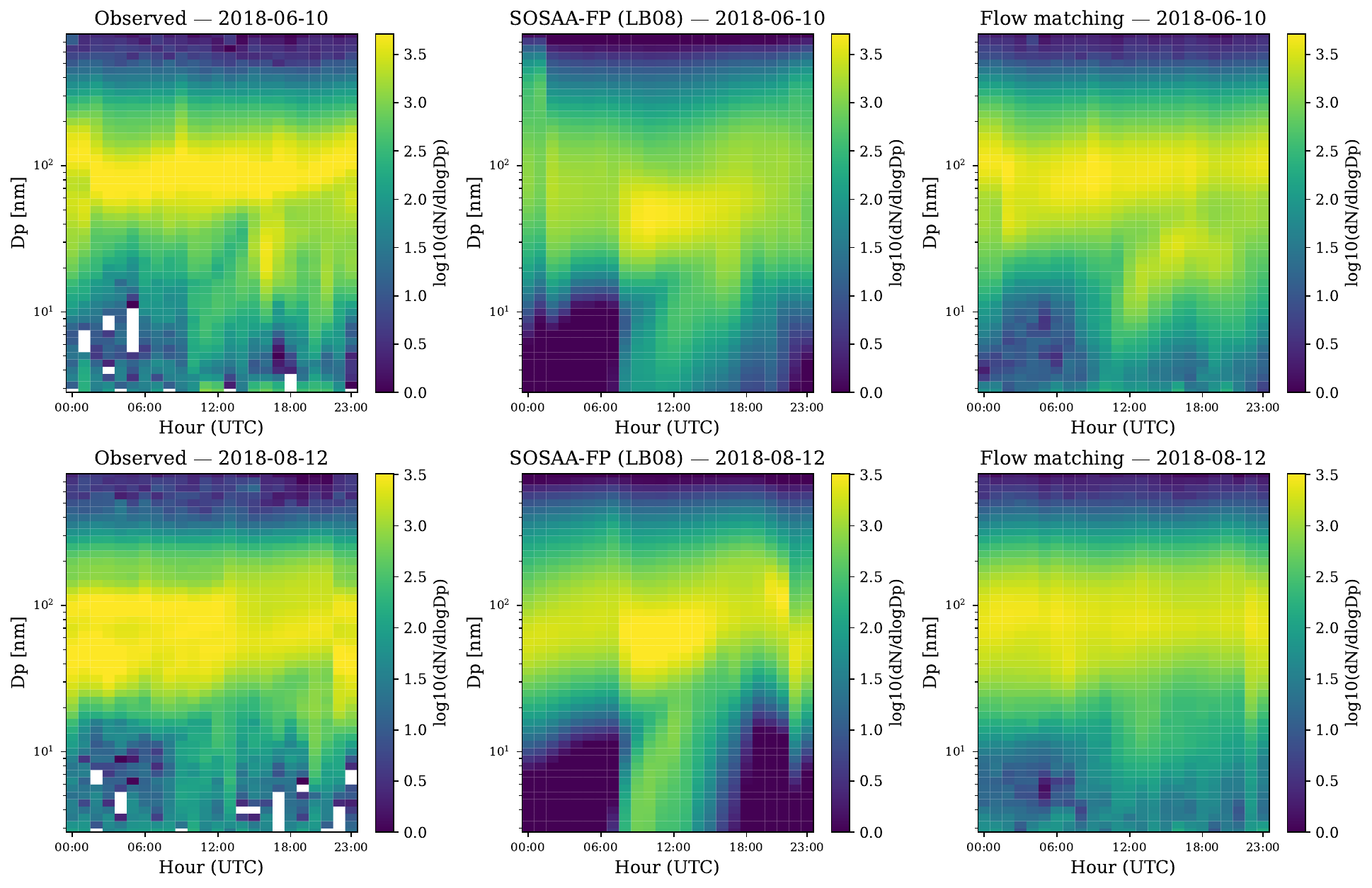}
\caption{\textbf{The same contrast holds through the day.} Time-resolved PNSDs for the cases of Fig.~\ref{fig:sosaa_examples}: (a--c)~10~June 2018 and (d--f)~12~August 2018. Within each row, panels show the observed DMPS PNSD, SOSAA-FP and Flow Matching. Color gives $\log_{10}(\mathrm{d}N/\mathrm{d}\log D_p)$ on a common scale within each row.}
\label{fig:sosaa_heatmaps}
\end{figure*}

On 645 matched windows, FM has a correlation of 0.938 and an RMSE of 1002\,cm$^{-3}$, against a correlation of 0.715 and an RMSE of 2105\,cm$^{-3}$ for SOSAA-FP (Table~\ref{tab:sosaa}). The MLP has the lowest RMSE (867\,cm$^{-3}$). In two matched cases, SOSAA-FP underestimates concentrations across most sizes by about an order of magnitude, whereas FM follows the observed amplitude above about 10\,nm (Figs.~\ref{fig:sosaa_examples} and \ref{fig:sosaa_heatmaps}). This comparison comes with several caveats. The data-driven models were fitted to local DMPS records, whereas SOSAA-FP is a simulation driven by emissions, chemistry and meteorology. It can therefore diagnose processes and sources that the learned models cannot. The result shows only that routine observations reproduce the local PNSD more closely in this matched setting. The running-mean and seasonal diagnostic of \citet{clusius2026} is given in Appendix~\ref{app:sosaa_seasonal}.


\begin{table*}[t]
\caption{\textbf{On matched 2018 windows at Hyyti\"al\"a, the learned models agree more closely with DMPS observations than SOSAA-FP does.} Metrics use 645 windows and all 50 common diameter bins (2.82--794\,nm); RMSE is in cm$^{-3}$. The ``vs SOSAA-FP'' rows measure disagreement between models and are not skill scores.}
\label{tab:sosaa}
\centering\scriptsize
\begin{tabular}{lcccccccc}
\tophline
Metric & \shortstack{SOSAA-\\FP} & \shortstack{Flow\\Matching} &
\shortstack{Diffusion\\Model} & \shortstack{FM +\\CFG} &
\shortstack{MLP} & \shortstack{FM +\\OT} &
\shortstack{Drifting\\Model} & \shortstack{Energy\\Matching} \\
\middlehline
\shortstack[l]{Correlation\\vs observed} & 0.715 & 0.938 & 0.868 & 0.924 & 0.957 & 0.950 & 0.849 & 0.913 \\
RMSE vs observed & 2105 & 1002 & 1531 & 1118 & 867 & 894 & 1378 & 1297 \\
\shortstack[l]{Correlation\\vs SOSAA-FP} & --- & 0.754 & 0.729 & 0.760 & 0.743 & 0.744 & 0.662 & 0.740 \\
RMSE vs SOSAA-FP & --- & 1763 & 971 & 1700 & 1865 & 1598 & 1317 & 1245 \\
\bottomhline
\end{tabular}
\belowtable{SOSAA-FP is a process-model simulation; the data-driven models were fitted to the local measurement record. Agreement with SOSAA-FP is therefore not a measure of accuracy.}
\end{table*}

\section{Discussion}
\label{sec:discussion}

\subsection{Interpretation}
In summary, conditioning a flow-matching model on PM$_{2.5}$ and meteorology yields PNSDs whose distribution, not only whose mean, tracks the observations, and a reduced GDE residual improves them further. Since this study is a benchmark on a two-station testbed, the value of these results lies less in the absolute scores than in what they reveal about model families, constraints and failure modes. Three results call for interpretation: the different verdicts of RMSE and CRPS, the contrast between constraint types, and the dependence of transfer on quantity and direction.

\textit{Why ensembles beat regression on CRPS but not always on RMSE.} Under the mass--number ambiguity, several distinct PNSDs are consistent with the same input. A squared-error regression returns their average, which RMSE rewards but which lacks the structure and spread of any single realization (Fig.~\ref{fig:realism}). A generative ensemble spreads probability across the plausible PNSDs, which CRPS rewards. The two metrics therefore answer different questions. Gradient boosting's lower RMSE at Hyyti\"al\"a and the generative models' lower CRPS at both sites are consistent: the former is the better point estimate at that site, the latter the better description of what remains uncertain.

\textit{Why the dynamical constraint helps and the integral constraints do not.} An integral constraint fixes one summary number and leaves many PNSDs that satisfy it. In this task, the mass reference also duplicates information already given as input, and because the resolved PNSD and the PM$_{2.5}$ instrument do not sample the same size range, forcing agreement can distort the PNSD. The GDE residual instead constrains how neighbouring bins and hours relate to each other, and so brings the strong autocorrelation of the PNSD in size and time into the training objective; it also widens the too-narrow FM ensemble. These observations are consistent with a regularizing effect of the dynamics, but the evidence is indirect: the constrained runs also have more parameters and a longer training budget.

\textit{Why transferability depends on quantity and direction.} \ncufp{} and CS weight the same PNSD differently. CS is weighted towards accumulation-mode particles, which tend to be transported regionally and are therefore shared between the two sites, whereas \ncufp{} is dominated by the smallest particles, which reflect more local sources, such as traffic at Kumpula and new particle formation at Hyyti\"al\"a. The median PNSDs in Fig.~\ref{fig:transfer_cost}b,c support this view: transferred models reproduce the accumulation mode in both directions and fail in the ultrafine range. We hypothesize that this explains why CS transfers better from Kumpula to Hyyti\"al\"a. The boreal-to-urban failure is consistent with a missing source: the Hyyti\"al\"a training record contains no analogue of the daytime UFP emissions from traffic at Kumpula, and a soft dynamical constraint cannot supply one. The present two-site comparison cannot separate the roles of emissions, formation, growth, transport and mixing, so these explanations remain hypotheses. A practical implication follows regardless: a PNSD model should be judged by the quantity and uncertainty that its application needs, since an average score can remain high while the relevant integral degrades.

\subsection{Limitations}
Several limitations follow from the deliberate choice of a small, controlled testbed, and the results should be read in that light. The evidence rests on two stations, one late-year holdout of about 60 days and one training run per configuration, so the bootstrap intervals do not include training-run variability; repeated training and holdouts from other seasons would address this. Two stations are enough to compare model families under identical conditions and to expose failure under transfer, but not to show how far neural networks can generalize when trained on many aerosol regimes. That question requires large, multi-station training data and is the natural next step. Every learned ensemble is too narrow, with 80\,\% intervals containing about half of the observations locally and fewer after transfer, and eight members limit how finely quantiles can be resolved. Post-hoc calibration, larger ensembles and guidance selected on validation data rather than the exploratory test-set sweep (Appendix~\ref{sec:res_cfg}) are natural remedies. The constants used to normalize the PNSDs may have been computed from the whole record rather than from the training period alone; if so, the models had indirect access to the range of values in the test period, which has not yet been ruled out. Instrument quality flags were also not used to remove questionable measurements (Appendix~\ref{app:data}). Finally, the analysis of sustained high-number episodes contains only 0, 3, 29 and 0 qualifying days in the four evaluation settings, too few for reliable verification of event probabilities (Appendix~\ref{sec:res_episode}).

The physical interpretation is also incomplete. The GDE gain cannot yet be attributed to the residual itself, because the constrained runs have an auxiliary head and a longer training budget; capacity- and budget-matched controls are needed. The growth rate is a particular weakness. Determining how fast particles grow, and what makes them grow, is a long-standing problem: even from direct DMPS measurements, growth rates can be determined reliably only for clear new-particle-formation events, and the results remain uncertain, partly because coagulation also shifts the PNSD. The sulfuric-acid proxy used here depends on SO$_2$, which is not among the stated inputs (Appendix~\ref{app:physics}), and is at best a weak indicator of growth, although it may indicate new particle formation. No proxy for the growth rate that transfers between environments has been established, yet the growth rate matters most for \ncufp. The integral constraints use references with different measurement support, and in this task the pseudo-mass reference is the input itself, so a size-resolved conversion between DMPS, PM$_{2.5}$ and CPC support would be required to make them physically meaningful. Finally, \ncufp{} and CS are computed from the same DMPS PNSDs used for evaluation and so do not validate process physics independently, and the SOSAA-FP comparison uses windows that overlap training data and a single model configuration, so it cannot rank data-driven and process-based modelling in general.

\conclusions
\label{sec:conclusions}
We asked how much of the aerosol PNSD can be recovered from the concurrent PM$_{2.5}$ and meteorological record. Treating this inverse problem as conditional generation, we benchmarked six generative models and three physical constraints on a controlled two-station testbed, with the aim of identifying which modelling choices matter rather than proposing a single best model. Four findings stand out. (1) Generative ensembles give better distributional skill than deterministic regression and historical means: all six generative models have lower CRPS than gradient boosting at both sites, although gradient boosting remains competitive on point error. (2) The form of a physical constraint matters: a reduced GDE lowers CRPS by 11.5\,\% and 5.5\,\% and widens the ensemble, whereas integral constraints on mass or total number alone degrade the generated PNSDs. (3) PM$_{2.5}$ and meteorology carry information on the PNSD beyond its seasonal and diurnal pattern: FM+GDE has the lowest CRPS of the learned models for \ncufp{} and CS at both sites, with skill scores of 0.27--0.83 against the historical ensemble, but its intervals contain only about half of the observations. (4) Transferability depends on the quantity and the direction: CS transfers better than \ncufp{} from the urban to the boreal site, whereas boreal-to-urban transfer fails for \ncufp, and in both directions the transfer error lies in the ultrafine range. Local PNSD estimation from routine observations is therefore feasible today, and the testbed has shown which ingredients are worth carrying forward.

These results raise the question of what would be required to estimate PNSDs at a station that measures only PM$_{2.5}$. First, and most importantly, the method should be scaled up: trained on large data sets from many stations that cover contrasting aerosol regimes, neural networks can be tested where their main strength lies, in learning from diverse data and generalizing to new domains and out-of-distribution conditions. The boreal-to-urban failure observed here, after training on a single station, is exactly the situation that multi-station training is designed to overcome. A natural intermediate test is a third background station with a long DMPS record, such as Hyltemossa in southern Sweden, which is less influenced by local traffic than Kumpula and more by transport from western Europe and the Atlantic than Hyyti\"al\"a; it would show at what point a boreal-trained model leaves its training distribution. Second, calibrated uncertainty is a prerequisite for any use of the intervals. Third, training-derived climatology could be represented explicitly, for example by starting generation from a historical PNSD template instead of pure noise. Fourth, source-sensitive predictors, such as NO$_x$, which is often measured at air-quality stations, or other traffic and emission indicators, could reduce the mass--number ambiguity directly. More generally, the approach could help to design measurement networks that balance cost against the value of the data: adding one or two relatively low-cost but informative predictors to an air-quality station could improve the inferred PNSD markedly, without investment in more advanced instruments such as a DMPS. Finally, capacity-matched ablations would clarify how much of the constraint's benefit is physical. These improvements are complementary to process models: data-driven estimates can extend the observational record, while process models remain necessary to explain it. The evaluation framework presented here can be used to measure progress along each of these axes.

\appendix

\section{Data preparation and evaluation splits}
\label{app:data}

\subsection{Size grid and data assembly}
The DMPS archives at both stations report 61 diameter bins between 1.00 and 1000\,nm. We use a fixed 50-bin subset between 2.82 and 794\,nm, excluding the nine smallest archived bins (1.00--2.51\,nm) and the two largest (891 and 1000\,nm), and we keep the archived bin centres without interpolation. Hourly values are means of the native 10\,min measurements, each weighted equally. ERA5 variables are taken from the grid cell nearest to each station. After assembly, each station has 61\,368 hourly records with complete PM$_{2.5}$, PNSD and ERA5 values on a common time axis.

\subsection{Transformations}
Particle concentrations span several orders of magnitude across bins and over time, so the model operates on log-transformed values. For each bin, the concentration $n$ is transformed to $\ell=\log(1+n)$ and then scaled linearly to the interval $[-1,1]$,
\begin{equation}
\tilde\ell=2\,\frac{\ell-\ell_{\min}}{\ell_{\max}-\ell_{\min}}-1 ,
\end{equation}
with fixed bin-wise constants $\ell_{\min}$ and $\ell_{\max}$ shared by both stations. The same scaling therefore applies at the training site and at the transfer target. Values outside the scaled range are kept rather than clipped. To return to physical units, the scaling is inverted and $n=\exp(\ell)-1$ is applied after negative values of $\ell$ are set to zero, which guarantees non-negative concentrations. PM$_{2.5}$ is transformed with $\log(1+x)$, and positive, strongly skewed meteorological variables are also log-transformed.

\subsection{Examples and splits}
An example is a 24\,h window of inputs together with the PNSD of the same 24 hours (Sect.~\ref{sec:data}). Windows start at every hour, so consecutive windows overlap by 23 hours, and a window is used only if all 24 hours are valid. Missing hours are kept as gaps on the time axis rather than removed, so that every window spans exactly 24 consecutive clock hours.

The data are split in time. All training windows end before 2 November 2024, 00:00\,UTC, and the temporal holdout consists of the 1417 windows that start on or after this cutoff (60 days, 2 November--31 December 2024). Before the cutoff, 10\,\% of the eligible windows (6048 per station), stratified by month and hour of day, form an \emph{interpolation} split, and the remaining 53\,857 windows per station are used for training. Because the interpolation windows overlap with neighbouring training windows by up to 23 hours, the interpolation split measures how well a model reproduces the training period, not its skill on unseen periods. We use it only for the diagnostics in Appendix~\ref{app:diagnostics} and for the matched comparison with SOSAA-FP (Sect.~\ref{sec:sosaa}); all skill statements in the main text refer to the temporal holdout. No separate validation set or early stopping is used.

\section{Generative models and baselines}
\label{app:models}

\subsection{Flow matching}
\begin{figure*}[htbp]
\includegraphics[width=\textwidth]{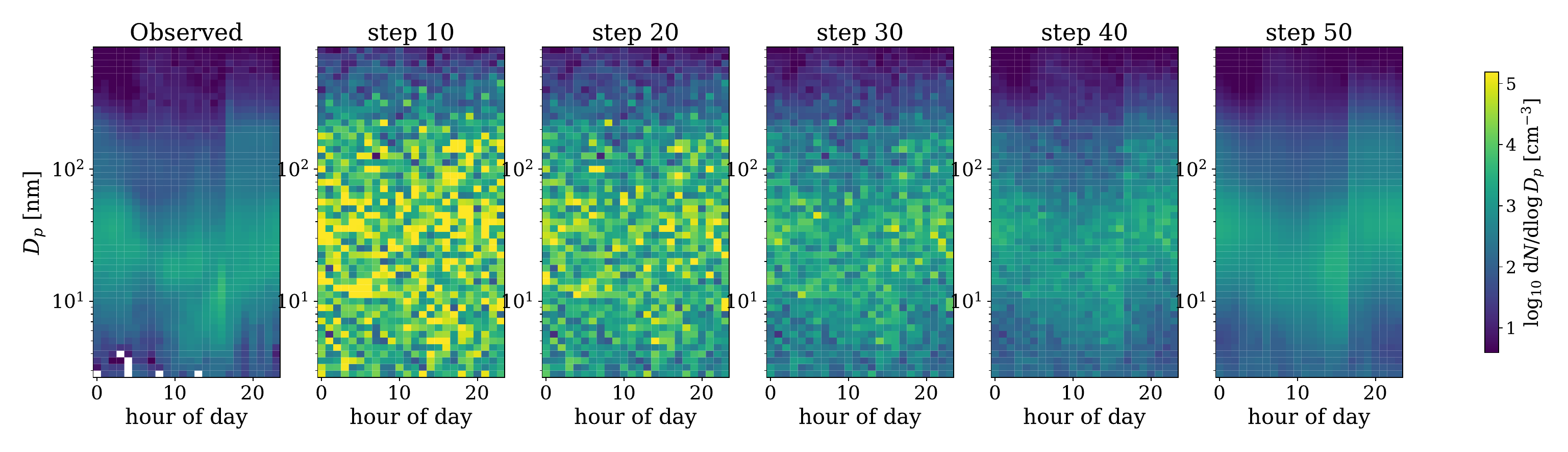}
\caption{\textbf{Flow Matching resolves Gaussian noise into a coherent 24\,h PNSD during integration.} Ensemble-mean state along the Euler integration of $\mathrm{d}\mathbf{x}/\mathrm{d}t=\mathbf{v}_\theta$ for one temporal-holdout sample (10~December 2024), from an almost purely Gaussian field at step~10 to the generated distribution at step~50, alongside the observation. Color is $\log_{10}\,\mathrm{d}N/\mathrm{d}\log D_p$; the vertical axis is mobility diameter.}
\label{fig:trajectory}
\end{figure*}

Flow matching connects a sample $\mathbf x_0$ of standard Gaussian noise and a data sample $\mathbf x_1$ by the straight path
\begin{equation}
\mathbf{x}_t = (1-t)\,\mathbf{x}_0 + t\,\mathbf{x}_1,
\qquad
\mathbf{v}^\ast = \frac{\mathrm d\mathbf x_t}{\mathrm dt}=\mathbf{x}_1 - \mathbf{x}_0 ,
\label{eq:interp}
\end{equation}
and trains the network $\mathbf v_\theta(\mathbf x_t,t,\mathbf c)$ to predict the velocity $\mathbf v^\ast$ from the intermediate state (Eq.~\ref{eq:fm}). Averaged over all pairs that pass through a given state, the regression target becomes the velocity of a transport that carries the whole noise distribution onto the conditional distribution of PNSDs \citep{lipman2023,liu2023}. A new sample is generated by integrating this velocity field from a fresh noise draw. We use the explicit Euler scheme with $K$ steps of size $\Delta t=1/K$,
\begin{equation}
\mathbf{x}^{(0)}\sim\mathcal{N}(0,I), \qquad
\mathbf{x}^{(k+1)} = \mathbf{x}^{(k)} + \Delta t\,
\mathbf{v}_\theta\!\left(\mathbf{x}^{(k)},t_k,\mathbf{c}\right),
\quad t_k=k\Delta t,\quad k=0,\ldots,K-1,
\label{eq:fm_sampling}
\end{equation}
with $K=50$ (Appendix~\ref{sec:appendix_nfe}). The integration is deterministic for a given starting point, so all variability within an ensemble comes from the different noise draws $\mathbf x^{(0)}$. Figure~\ref{fig:trajectory} shows how an initially random field is resolved into a coherent PNSD along this path.

\subsection{Network architecture}
The velocity network is a Transformer \citep{vaswani2017} that represents a 24\,h window as a sequence of 24 hourly elements. For each hour, the current 50-bin state $\mathbf x_t$ and the inputs of that hour are projected into a common feature space and added, and a learned position embedding identifies the hour within the window. Self-attention layers then exchange information between all hours. The generative time $t$ enters through adaptive layer normalization, which rescales the features of every layer according to $t$; the network can therefore behave differently near the noise end and near the data end of the path. A final projection maps each hour back to 50 bins. The network has about $1.24\times10^6$ trainable parameters. The diffusion model, the classifier-free-guidance variant and the optimal-transport variant use the same network, so their differences reflect the generative objective rather than the architecture. Training settings are provided with the code (see Code availability); for interpreting the comparisons, the relevant difference is that plain FM, diffusion, FM+CFG and the MLP were trained for 200 epochs, whereas FM+OT, the Drifting Model, Energy Matching and all constrained FM variants were trained for 400 epochs.

\subsection{Comparison generative models}
\textit{Diffusion.} A denoising diffusion model \citep{ho2020} corrupts a data sample $\mathbf y$ gradually with Gaussian noise, $\mathbf x_s=\sqrt{\bar\alpha_s}\,\mathbf y+\sqrt{1-\bar\alpha_s}\,\boldsymbol\epsilon$, where $\bar\alpha_s$ decreases from one to nearly zero over the noising steps $s$ following a linear schedule. The network is trained to predict the added noise $\boldsymbol\epsilon$, and samples are generated by reversing the corruption step by step, adding fresh noise at each step. Unlike FM, the sampling path is stochastic.

\textit{Optimal-transport coupling (FM+OT).} Standard FM pairs noise and data samples at random, so the straight paths of different pairs can cross. FM+OT instead pairs the samples within each training batch by solving an assignment problem that minimizes the total squared distance between paired samples \citep{tong2024}. The resulting paths are shorter and cross less often, which can make the learned velocity field smoother.

\textit{Classifier-free guidance (FM+CFG).} During training, the conditioning $\mathbf c$ is replaced by a learned null input $\emptyset$ for 15\,\% of the examples, so that one network learns both the conditional and the unconditional velocity \citep{ho2022}. At sampling time, the two are combined as
\begin{equation}
\tilde{\mathbf v}=\mathbf v_\theta(\mathbf x_t,t,\emptyset)+ w\left[\mathbf v_\theta(\mathbf x_t,t,\mathbf c)- \mathbf v_\theta(\mathbf x_t,t,\emptyset)\right].
\end{equation}
A guidance scale $w=1$ recovers ordinary conditional sampling, $w>1$ strengthens the influence of the inputs, and $w<1$ moves the samples towards the unconditional distribution, which widens the ensemble. The tables use $w=1$; the effect of $w$ is shown in Appendix~\ref{sec:res_cfg}.

\textit{Drifting Model.} The Drifting Model \citep{deng2026} learns a generator that maps noise to a sample in a single network evaluation. During training, the distribution of generated samples is moved step by step towards the data distribution, so that no integration is needed at inference.

\textit{Energy Matching.} Energy Matching \citep{balcerak2025} learns a scalar potential whose negative gradient defines the transport from noise to data. Because the potential does not depend on the generative time, the network does not receive $t$ as an input. We train it with its flow-based objective only; the additional contrastive refinement stage of the method was not used, because it was unstable on these data.

\subsection{Deterministic baselines and non-learning references}
Gradient boosting fits one regression model per diameter bin. Each hour of a training window is treated as a separate sample, with the hour of day encoded by its sine and cosine, so the model can adapt its estimate to the time of day but does not represent dependence between bins or between hours. The MLP is a fully connected network that maps the complete 24\,h input window to all $24\times50$ outputs at once. Both are trained on squared error and return a single PNSD per input.

The month/hour climatology is the mean training-period PNSD for each of the 288 combinations of calendar month and hour of day; each target hour is assigned the value of its own month and hour. The seasonal template is defined in the same way using seasons instead of months. The historical ensemble consists of eight complete 24\,h PNSD sequences drawn from the training period of the source station, each starting in the same season and at the same hour of day as the target window. Because whole sequences are used, every member retains the observed dependence across bins and hours. The historical ensemble ignores the current inputs and therefore shows what can be estimated from the training archive alone.

\section{Physical constraints}
\label{app:physics}

\subsection{Resolved mass and number}
\label{sec:mass}\label{sec:ncc}
The resolved mass and number (Eqs.~\ref{eq:mass} and \ref{eq:l_cn}) are sums over the 50 bins of the predicted PNSD, using logarithmic bin widths $\Delta_i=\Delta\log_{10}D_{p,i}$ and a constant particle density $\rho=1.5$\,g\,cm$^{-3}$. PM$_{2.5}$ is defined by the aerodynamic diameter, whereas the DMPS sizes particles by their electrical mobility. For spherical particles, the two diameters are approximately related by $D_{\rm m}\approx D_{\rm a}(\rho/\rho_0)^{-1/2}$ with $\rho_0=1$\,g\,cm$^{-3}$, neglecting shape and slip corrections. The PM$_{2.5}$ cut therefore corresponds to a mobility diameter of about $2.5\,\mu\mathrm m\times1.5^{-1/2}\approx2.0\,\mu$m, well above the 794\,nm upper limit of the grid. All predicted bins thus lie inside the PM$_{2.5}$ size range, but the mass between 794\,nm and about 2\,$\mu$m is missing from the resolved mass, which is why the mass term is used only as a soft reference.

Both integral constraints compare logarithms of the resolved and the measured quantity through the Huber function,
\begin{equation}
\operatorname{Huber}_{\delta}(u)=
\begin{cases}
u^2/2, & |u|\leq\delta,\\
\delta(|u|-\delta/2), & |u|>\delta,
\end{cases}
\qquad \delta=0.5 ,
\label{eq:huber}
\end{equation}
where $u$ is the difference of the two logarithms. In logarithmic space, a constant multiplicative mismatch, such as the missing mass above 794\,nm, becomes a constant offset, and the linear branch of the Huber function prevents such systematic offsets from dominating the gradient. The reference scales $M_0$ and $N_0$ in Eqs.~(\ref{eq:l_mass}) and (\ref{eq:l_cn}) make the logarithms well defined at zero concentration, and negative reference values, which can arise from instrument noise, are set to zero. The \nctot{} constraint is evaluated only in hours with a valid CPC measurement.

Each physical term is weighted so that its initial magnitude is comparable to that of the FM loss; for the mass term, this gives $\lambda_{\rm m}=0.05$, and the number and GDE weights were set by the same rule. A mass weight of 1.0 led to a degenerate solution that satisfied the mass constraint at the expense of a realistic PNSD, which illustrates why the mass term must remain soft.

The CPC and the DMPS measure closely related but not identical quantities. Over the full record, their log-transformed total number concentrations correlate with Pearson $r=0.988$ at Hyyti\"al\"a (59\,689 hours with both measurements) and $r=0.963$ at Kumpula (57\,424 hours), with median DMPS/CPC ratios of 0.87 and 1.03, respectively. Possible contributors to the differences include particles below the DMPS lower limit, which the CPC counts, and differences in detection efficiency and sampling losses \citep{wiedensohler2012}, which the number constraint does not model.

\subsection{Condensation and coagulation sinks}
\label{sec:sinks}
The condensation sink (CS; Eq.~\ref{eq:cs}) is the rate at which a condensable vapor is lost to the particle population \citep{kulmala2001}. The Fuchs--Sutugin correction accounts for the transition between the continuum regime, where vapor diffuses to large particles, and the free-molecular regime of small particles,
\begin{equation}
\beta_m(D_p)=\frac{1+\mathrm{Kn}}{1+\left(\frac{4}{3\alpha}+0.377\right)\mathrm{Kn}+\frac{4}{3\alpha}\mathrm{Kn}^2},
\qquad \mathrm{Kn}=\frac{2\lambda_v}{D_p},
\end{equation}
where $\lambda_v$ is the mean free path of the vapor molecules and $\alpha$ the accommodation coefficient, set to one. For small particles, $\beta_m D_p$ grows approximately with $D_p^2$, and for large particles approximately with $D_p$, so the sink is weighted towards larger particles. The coagulation sink of a particle of diameter $d$,
\begin{equation}
\mathrm{CoagS}(d) = \sum_i K(d, D_{p,i})\,N_i ,
\label{eq:coags}
\end{equation}
is its loss rate by coagulation with the existing particles, where $K$ is the Brownian coagulation kernel \citep{kulmala2001,seinfeld2016} and $N_i=n_i\Delta\log_{10}D_{p,i}$ is the number concentration in bin $i$. Both sinks are evaluated at fixed reference conditions (293.15\,K, 101325\,Pa) for predicted and observed PNSDs alike, so that differences between them reflect the size distribution only.

\subsection{Reduced general dynamic equation}
\label{sec:gde}
For a size distribution $n=\mathrm dN/\mathrm d\log_{10}D_p$ at a fixed station, the continuity equation underlying the GDE constraint is
\begin{equation}
\frac{\partial n}{\partial\tau}
+\frac{\partial}{\partial\ln D_p}
\left[\frac{\mathrm{GR}}{D_p}n\right]
=S-\mathrm{CoagS}\,n-\lambda_{\mathrm{dil}}n ,
\label{eq:gde}
\end{equation}
where $\tau$ is physical (clock) time. The second term on the left describes condensational growth as a flux through size space with velocity $\mathrm{GR}/D_p$ in $\ln D_p$. On the right, $S$ collects sources and transport, the coagulation term removes particles that collide with the existing population, and $\lambda_{\mathrm{dil}}=(\mathrm dH/\mathrm d\tau)_+/H$ dilutes the population when the boundary layer of height $H$ grows. The equation is reduced in three respects: coagulation enters only as a loss, not as a gain at larger sizes; deposition is not represented explicitly; and new particle formation appears only as the growth flux into the smallest resolved bin. The source term $S$ is unconstrained in sign and can absorb these omitted processes, as well as the advection of different air masses.

The residual of Eq.~(\ref{eq:gde}) is evaluated on the predicted PNSD (Eq.~\ref{eq:gde_residual}). Time derivatives are computed by finite differences over the 24 hourly values, central in the interior and one-sided at the first and last hour, and size derivatives by upwind differences in $\ln D_p$, which follow the direction of growth. At the lower edge of the grid, the incoming flux is set by the growth velocity of the first bin; it represents growth into the observed range rather than a nucleation rate. All rates are expressed per hour. The growth rate and the source term are predicted by an auxiliary network head from the conditioning only, so that they do not inherit noise from the intermediate generative state. The growth rate is constrained to be non-negative and is informed by a sulfuric-acid proxy proportional to SO$_2$ concentration times radiation divided by CS \citep{petaja2009}; the dependence of this proxy on SO$_2$, which is not among the model inputs, is discussed in Sect.~\ref{sec:discussion}. The head also predicts the log residual variance $s_{h,i}$ of Eq.~(\ref{eq:l_gde}), which is bounded to $[-5,5]$, i.e. to variances between $6.7\times10^{-3}$ and $1.5\times10^{2}$, so that the model can neither ignore a residual entirely nor overweight it without limit. Because $S$ and $s_{h,i}$ are flexible, a low residual shows consistency with Eq.~(\ref{eq:gde}) but does not uniquely identify the growth or source rates.

\subsection{Coupling the constraints to flow matching}
\label{sec:coupling}
During training, the network sees partly noised states $\mathbf x_t$, whereas the physical relationships apply to clean PNSDs. Inverting the straight path of Eq.~(\ref{eq:interp}) gives the clean-sample estimate implied by the current velocity prediction,
\begin{equation}
\hat{\mathbf{x}}_1 = \mathbf{x}_t + (1-t)\,\mathbf{v}_\theta ,
\label{eq:x1hat}
\end{equation}
which equals $\mathbf x_1$ exactly when the predicted velocity is correct. The constraints are evaluated on $\hat{\mathbf x}_1$ after conversion to physical units, and their gradients flow back into $\mathbf v_\theta$. Near the noise end ($t\to0$), $\hat{\mathbf x}_1$ is an extrapolation from an almost random state, so the weights $t^2$ and $t^3$ of Eq.~(\ref{eq:total}) suppress the constraints there; the GDE term, which contains derivatives that amplify noise, receives the steeper weight. The constraints act only on the training objective; generation at inference proceeds exactly as for unconstrained FM. All physical ablations use FM as the base model, and every constrained variant uses this time weighting.

\section{Evaluation metrics and diagnostics}
\label{app:metrics}

\subsection{Scores and their aggregation}
For generative models, point scores (RMSE and correlation) use the ensemble mean. The correlation is computed for each window on the log-transformed PNSD, treating its $50\times24$ values as one vector, and then averaged over windows. It measures agreement in the shape of the PNSD and in its temporal evolution, and it can be high simply because the typical shape of the PNSD persists from window to window.

The CRPS of Eq.~(\ref{eq:crps}) uses the fair estimator for a finite ensemble of $M=8$ members, which corrects the tendency of the ordinary estimator to penalize small ensembles; for deterministic predictions ($M=1$), the pairwise term vanishes and the CRPS reduces to the absolute error. The CRPS scores each bin and hour separately and therefore does not test whether the dependence between bins and hours is realistic.

Coverage is computed from the empirical 10th and 90th percentiles of the eight members. With eight members, these percentiles are coarse, so coverage values should be read as approximate. Coverage can be aggregated over bins in two ways: Table~\ref{tab:variance} reports the median of the per-bin coverage (0.454 for FM), whereas the guidance sweep in Appendix~\ref{sec:res_cfg} reports the mean (0.528 for FM). The rank of an observation within its ensemble counts the members that lie strictly below it; for a calibrated ensemble, all ranks are equally likely. The spread--skill ratio is the mean ensemble standard deviation divided by the RMSE of the ensemble mean; values near one indicate an ensemble whose spread matches its error, and values below one indicate an ensemble that is too narrow.

Confidence intervals for the mean CRPS use a block bootstrap with 1000 replicates. Because consecutive windows overlap, the data are resampled in blocks of 24 consecutive windows, and the 2.5th and 97.5th percentiles of the resampled means form the 95\,\% interval. These intervals describe sampling uncertainty over windows, not variation between training runs. Ensemble members are drawn anew in each evaluation, so two evaluations of the same trained model differ slightly, as for FM in Tables~\ref{tab:performance} and \ref{tab:physics_indomain} (CRPS of 142 and 141.9\,cm$^{-3}$).

\subsection{Interpolation, variability and calibration}
\label{app:diagnostics}
Three diagnostics describe how much of the observed variability a model reproduces. The standard-deviation ratio compares the standard deviation of the predicted and observed PNSDs across windows. The texture ratio compares their bin-to-bin variability within a window, which measures how much fine structure across diameter bins is retained. The temporal correlation is the correlation between the predicted and observed time series of total number concentration. For generative models, all three use the ensemble mean, which is expected to be smoother than any single observation, so ratios below one partly reflect averaging rather than model deficiency.

\begin{figure*}[htbp]
\includegraphics[width=\textwidth]{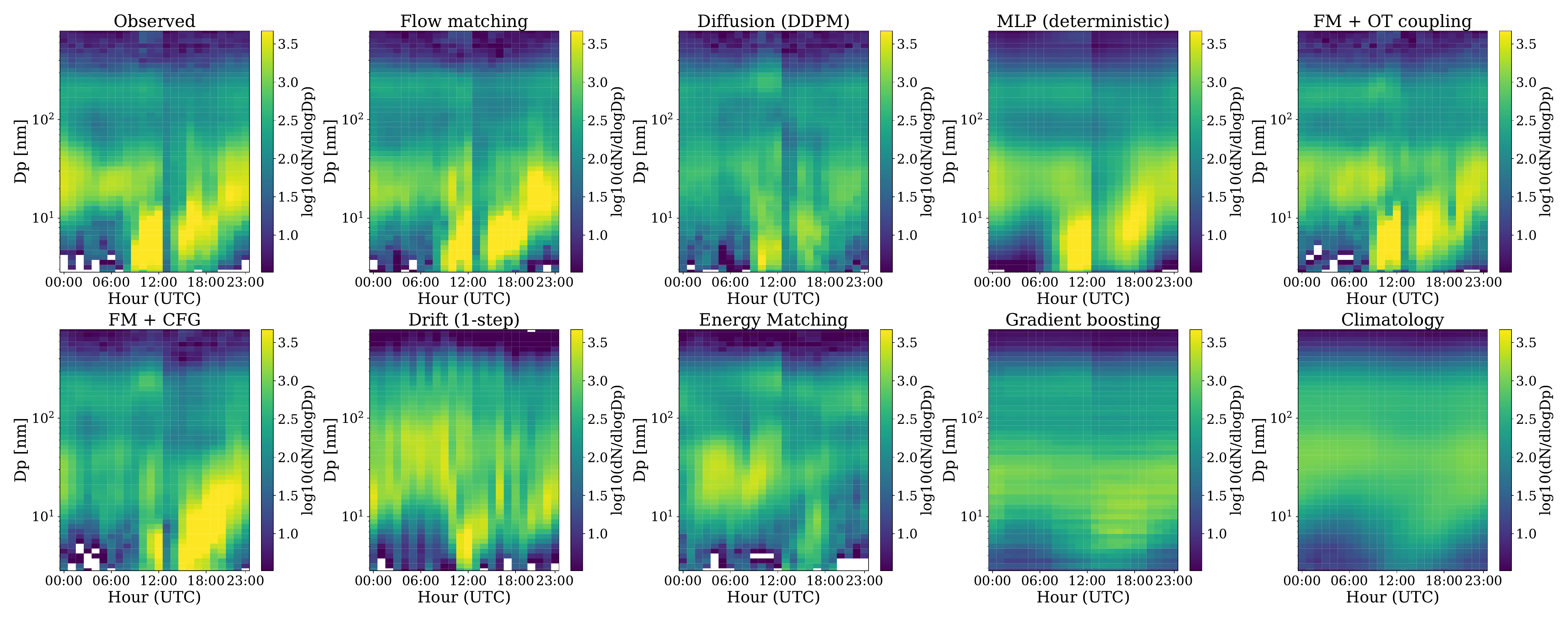}
\caption{\textbf{On the overlapping interpolation split, all learned models reproduce the observed structure closely.} Illustrative Hyyti\"al\"a example. The first panel is the observation; subsequent panels show one draw from each generative model or the deterministic prediction. Color is $\log_{10}\,\mathrm{d}N/\mathrm{d}\log D_p$; the vertical axis is mobility diameter and the horizontal axis is UTC hour.}
\label{fig:grid_interp}
\end{figure*}

\begin{figure*}[htbp]
\includegraphics[width=0.9\textwidth]{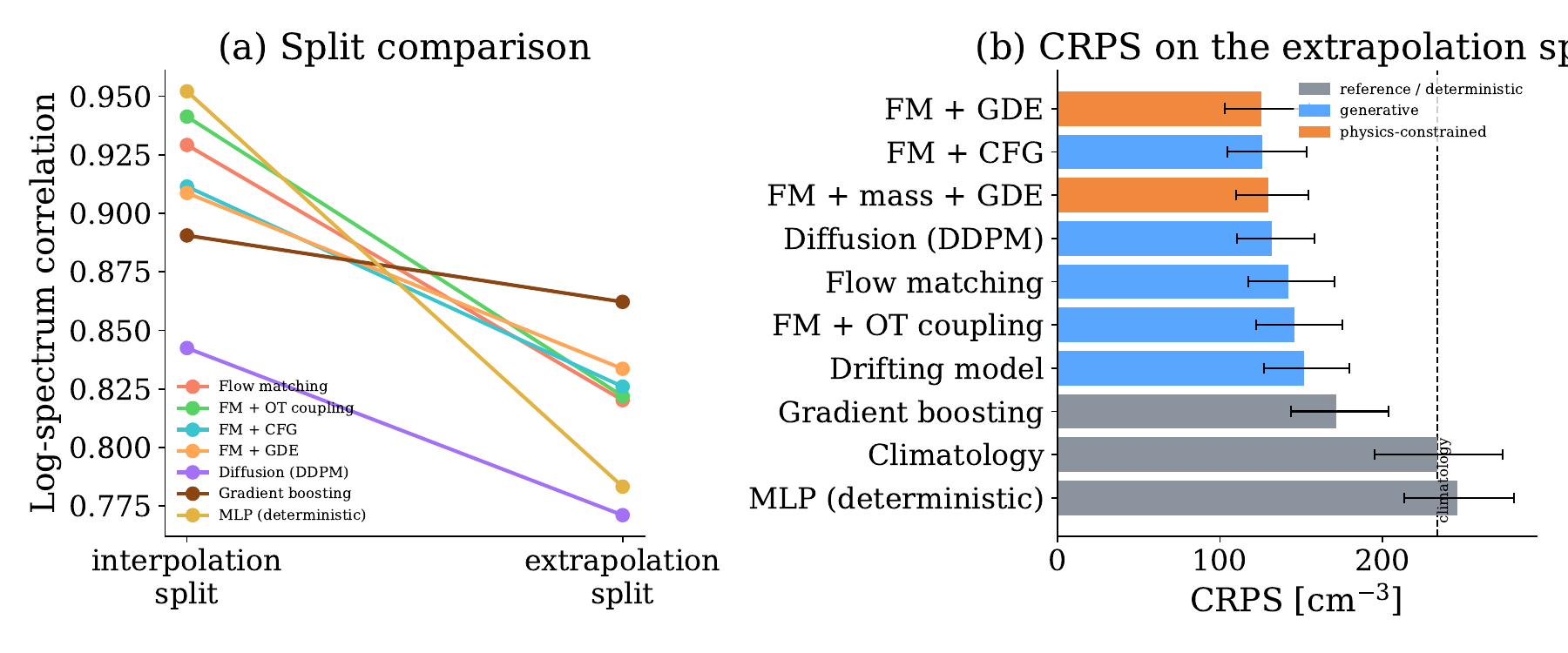}
\caption{\textbf{Scores on the overlapping interpolation split are not estimates of temporal skill.} Hyyti\"al\"a. (a) Mean correlation of log-transformed PNSDs on the interpolation split and temporal holdout; lines connect the same model. (b) Temporal-holdout CRPS with 95\,\% block-bootstrap intervals; the dashed line denotes climatology.}
\label{fig:performance}
\end{figure*}

\begin{figure*}[htbp]
\includegraphics[width=\textwidth]{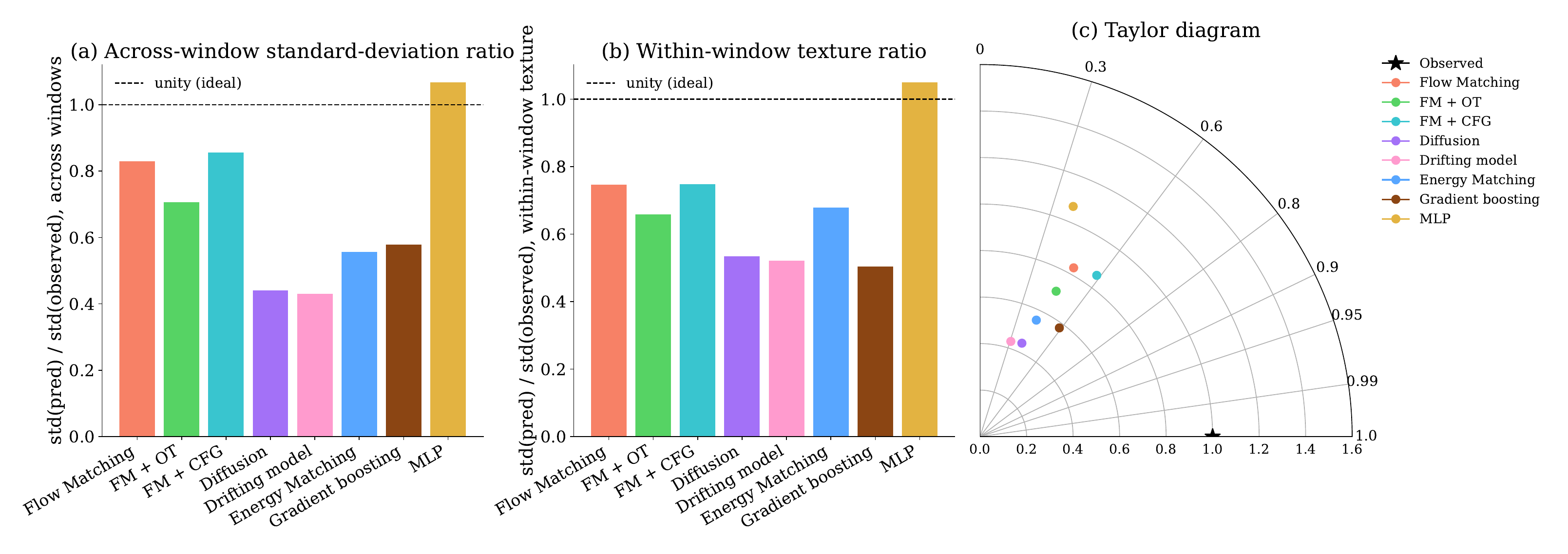}
\caption{\textbf{Flow Matching preserves more of the observed PNSD variability than gradient boosting.} Temporal holdout, Hyyti\"al\"a. (a)~Ratio of predicted to observed standard deviation across windows and (b)~ratio of predicted to observed within-window texture across diameter bins; unity is ideal. (c)~Taylor diagram combining the standard-deviation ratio (radius) with temporal correlation (angle); the black star marks the observation. Generative models use the ensemble mean.}
\label{fig:variance}
\end{figure*}

\begin{figure*}[htbp]
\includegraphics[width=\textwidth]{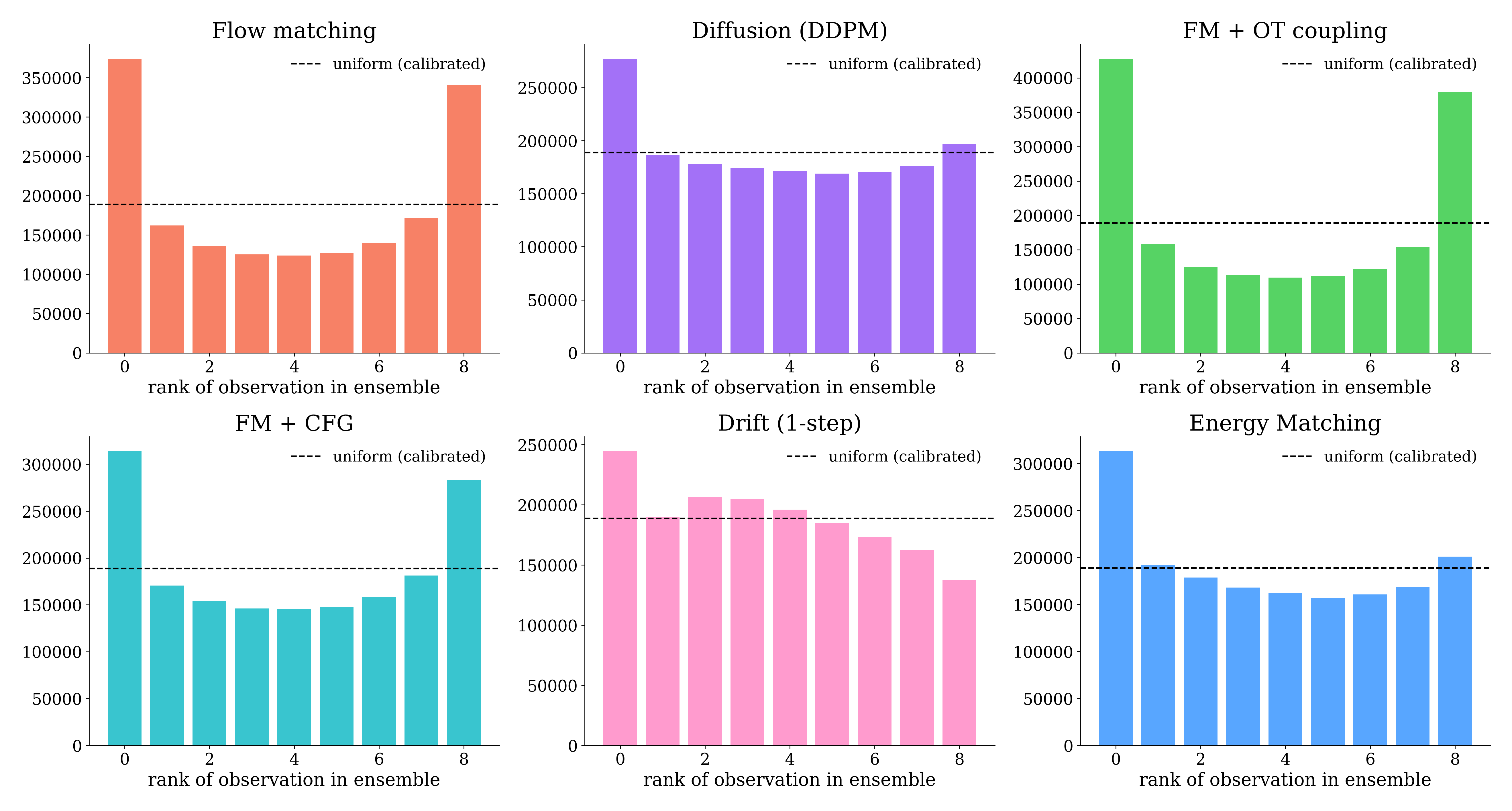}
\caption{\textbf{No generative ensemble is calibrated.} Ensemble rank histograms on the temporal holdout. Uniform ranks are expected for exchangeable observations and members (dashed line). A U-shape can reflect under-dispersion or a mixture of conditional biases; a dome indicates over-dispersion. Coverage is given in Table~\ref{tab:variance}.}
\label{fig:rank}
\end{figure*}

On the overlapping interpolation split, the MLP reaches a correlation of 0.95, a value that reflects the overlap with training windows rather than skill on new periods (Figs.~\ref{fig:grid_interp} and \ref{fig:performance}). On the temporal holdout, FM retains standard-deviation and texture ratios of 0.83 and 0.75, compared with 0.58 and 0.50 for gradient boosting (Table~\ref{tab:variance}, Fig.~\ref{fig:variance}). The rank histograms of all generative models depart from uniformity (Fig.~\ref{fig:rank}), indicating ensembles that are too narrow, conditionally biased, or both.

\begin{figure*}[htbp]
\includegraphics[width=\textwidth]{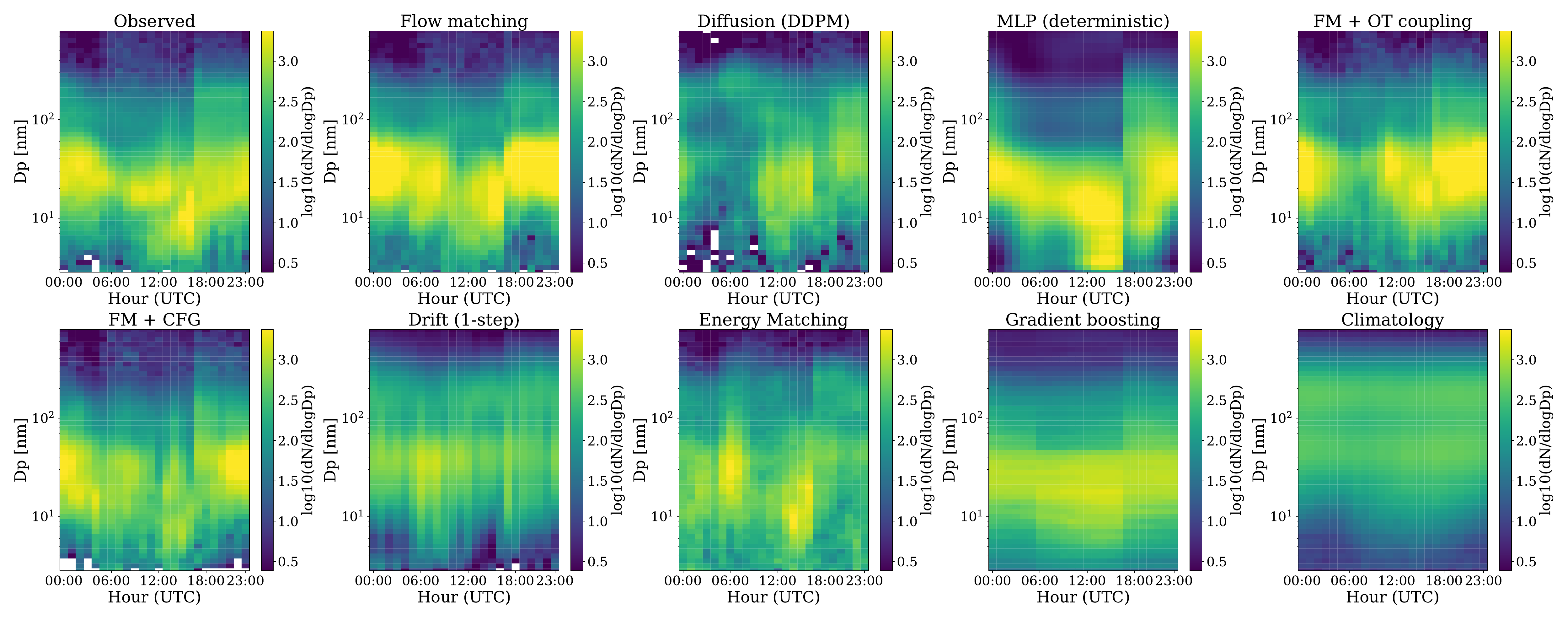}
\caption{\textbf{Generative models reproduce hour-to-hour structure of the PNSD that deterministic estimates and the climatology smooth out.} Generated 24\,h PNSDs for one temporal-holdout day at Hyyti\"al\"a, compared with the observation (top left). Each generative panel shows a single ensemble member. Color is $\log_{10}\,\mathrm{d}N/\mathrm{d}\log D_p$; the vertical axis is mobility diameter and the horizontal axis is UTC hour.}
\label{fig:grid_extrap}
\end{figure*}

Figure~\ref{fig:grid_extrap} complements Fig.~\ref{fig:realism} by showing one generated PNSD from every model for one temporal-holdout day at Hyyti\"al\"a. The FM variants reproduce the hour-to-hour movement of the Aitken mode. Gradient boosting and the climatology return smooth bands that barely change through the day, the Drifting Model shows hour-to-hour striping, and the MLP reproduces the structure but exaggerates the nucleation-mode band. The abrupt change in the observed PNSD in the late afternoon may be an instrument artefact or a remnant of earlier gap filling in the archived data, so models that reproduce it should not be credited with skill for that feature.

\begin{table}[htbp]
\caption{Distribution diagnostics for the Hyyti\"al\"a temporal holdout. Variability ratios use the ensemble mean. Coverage is the median across bins of central-80\,\% interval coverage (nominal 0.80); definitions are in Appendix~\ref{app:metrics}.}
\label{tab:variance}
\centering\small
\begin{tabular}{lrrrr}
\tophline
Model & std. ratio & texture ratio & coverage & temporal corr. \\
\middlehline
Flow Matching & 0.83 & 0.75 & 0.45 & 0.48 \\
FM + OT & 0.71 & 0.66 & 0.46 & 0.41 \\
FM + CFG & 0.86 & 0.75 & 0.59 & 0.51 \\
Diffusion Model & 0.44 & 0.53 & 0.41 & 0.58 \\
MLP & 1.07 & 1.05 & n/a & 0.00 \\
Gradient Boosting & 0.58 & 0.50 & n/a & 0.00 \\
Drifting Model & 0.43 & 0.52 & 0.65 & 0.31 \\
\bottomhline
\end{tabular}
\end{table}

\subsection{Generative objectives and small-particle errors}
\label{sec:res_objectives}
The generative objectives trade different aspects of skill. At Hyyti\"al\"a, optimal-transport coupling slightly improves the correlation (from 0.820 to 0.822) and the RMSE (from 389\,cm$^{-3}$ to 383\,cm$^{-3}$) of FM but worsens its CRPS (from 142\,cm$^{-3}$ to 146\,cm$^{-3}$), and the one-step Drifting Model reaches a CRPS of 152\,cm$^{-3}$ and an RMSE of 425\,cm$^{-3}$ at a fraction of the sampling cost. No objective is best in accuracy, coverage and cost at once.

\begin{figure}[htbp]
\includegraphics[width=0.9\textwidth]{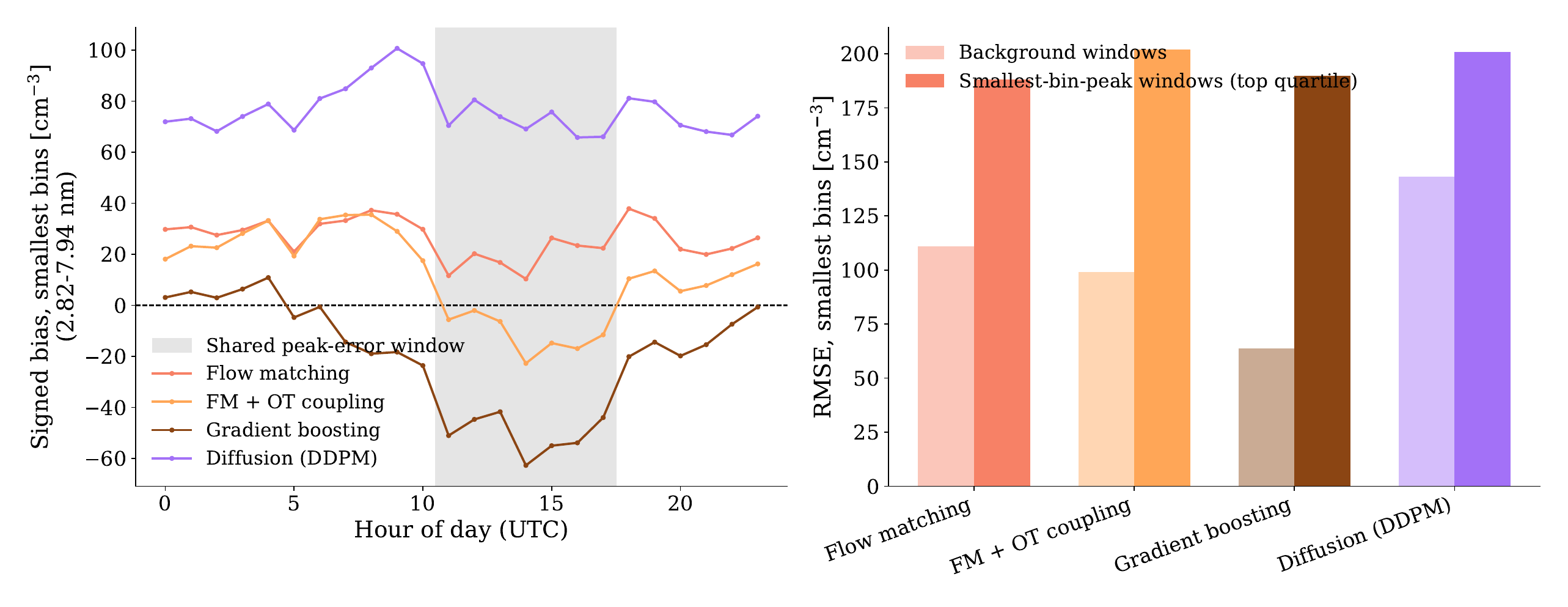}
\caption{\textbf{Errors in the smallest bins concentrate around midday for every model.} Exploratory diagnostics for the ten smallest resolved bins (2.82--7.94\,nm) at Hyyti\"al\"a. (a) Signed bias by UTC hour; the 11--17\,UTC window was selected by inspecting these errors. (b) Background and peak-window RMSE; peak windows reach at least the holdout 75th-percentile amplitude ($n=356$ of 1417). Values are in Table~\ref{tab:diagnostics}.}
\label{fig:diurnal}
\end{figure}

Errors in the smallest bins are of particular interest because these bins carry the signature of new particle formation. We therefore examine the ten smallest bins (2.82--7.94\,nm) separately (Fig.~\ref{fig:diurnal}, Table~\ref{tab:diagnostics}). Windows whose maximum observed concentration in these bins reaches at least the 75th percentile of the holdout form a peak group (356 of 1417 windows, 25.1\,\%); the remaining windows form the background group. Errors concentrate between 11 and 17\,UTC, a period selected by inspecting the holdout errors, so this analysis is descriptive. FM has the lowest RMSE in the peak group (188.1\,cm$^{-3}$, against 189.9\,cm$^{-3}$ for gradient boosting), whereas gradient boosting has the lowest RMSE in the background group (63.6\,cm$^{-3}$). The peak group is defined by concentration alone and does not identify new particle formation events.

\begin{table}[htbp]
\caption{Diurnal and smallest-bin peak-window errors at Hyyti\"al\"a for the ten smallest resolved bins (2.82--7.94\,nm), averaged over 11--17\,UTC. Peak windows are the top quartile by smallest-bin amplitude ($n=356$ of 1417). Bias and RMSE are in cm$^{-3}$.}
\label{tab:diagnostics}
\centering
\scriptsize
\begin{tabular}{lrrrr}
\tophline
Model & bias & bg. RMSE & peak RMSE & ratio \\
\middlehline
Flow Matching & 18.8 & 111.0 & 188.1 & 1.69 \\
FM + OT & -11.4 & 99.0 & 202.1 & 2.04 \\
Gradient Boosting & -50.4 & 63.6 & 189.9 & 2.98 \\
Diffusion Model & 71.7 & 143.2 & 200.8 & 1.40 \\
\bottomhline
\end{tabular}
\end{table}

\subsection{Guidance trade-off}
\label{sec:res_cfg}
\begin{figure}[htbp]
\includegraphics[width=0.9\textwidth]{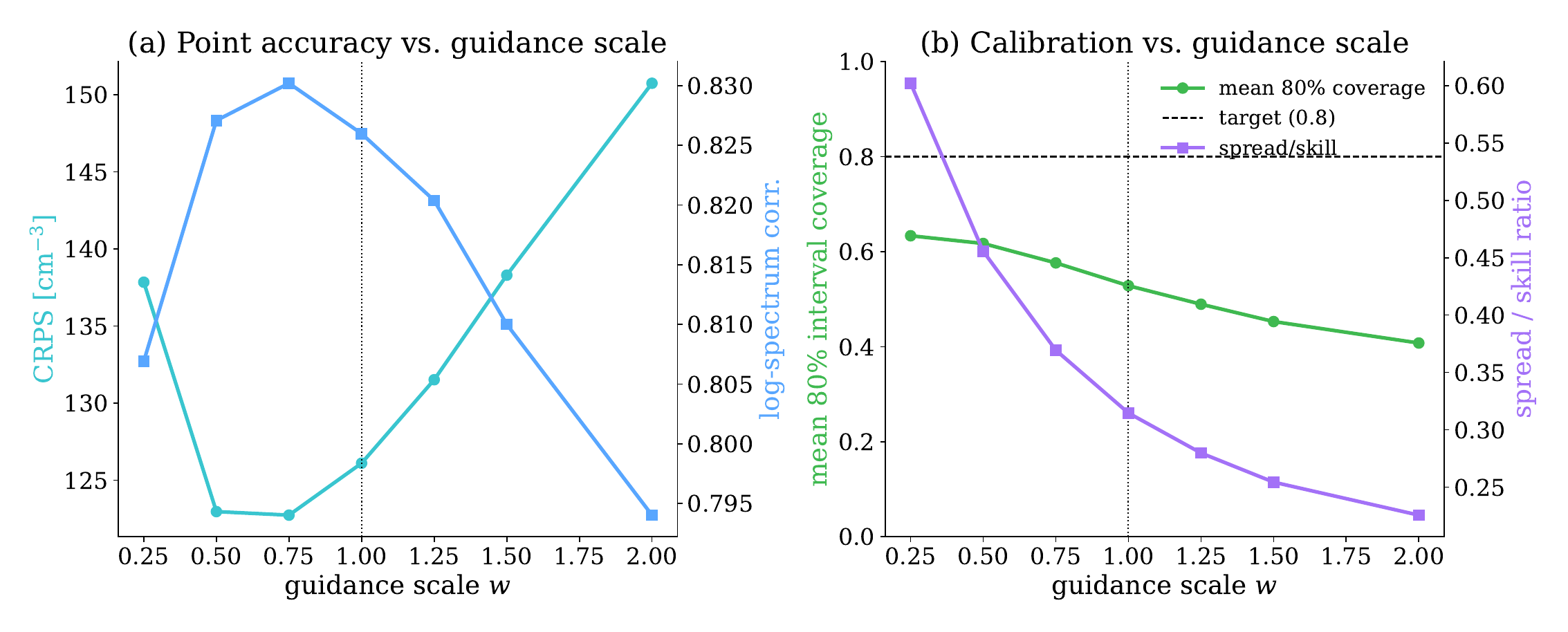}
\caption{\textbf{Guidance scale trades CRPS against interval coverage.} Classifier-free-guidance sweep for FM+CFG on the Hyyti\"al\"a temporal holdout. (a)~CRPS and correlation; (b)~mean 80\,\% interval coverage and spread/skill ratio against the guidance scale $w$. $w=1$ (dotted line) is plain conditional sampling; $w<1$ blends towards the unconditional distribution.}
\label{fig:cfg}
\end{figure}

The guidance scale $w$ of FM+CFG controls how strongly the samples follow the inputs, and therefore how wide the ensemble is (Fig.~\ref{fig:cfg}). At $w=1$, FM+CFG improves on plain FM (CRPS of 126\,cm$^{-3}$ against 142\,cm$^{-3}$) but remains too narrow, with a mean coverage of 0.53. Reducing $w$ to 0.25 widens the ensemble, raising the coverage to 0.63 and the spread--skill ratio to 0.60, at the cost of a higher CRPS (138\,cm$^{-3}$). Increasing $w$ to 2 narrows it, and the coverage drops to 0.41. The lowest CRPS of the sweep, 123\,cm$^{-3}$, occurs at $w=0.75$ and is practically equal to that at $w=0.5$. Because the sweep was evaluated on the holdout, it was not used to choose the reported operating point $w=1$; selecting $w$ in an application would require separate validation data.

\section{Extended spatial transfer}
\label{app:transfer}
The paired tests in Sect.~\ref{sec:res_ood} score transferred and locally trained models on the same 60-day holdout. As a complementary test over a longer period, Table~\ref{tab:crossdomain} applies each model trained at one station, without retraining, to all valid windows at the other station. Because these windows cover a different period from the local holdouts, the scores in this appendix should not be compared directly with those in Table~\ref{tab:performance}. Absolute errors are larger at Kumpula partly because its concentrations are higher.

In this longer test, diffusion has the highest correlation in both directions (0.792 at Kumpula and 0.689 at Hyyti\"al\"a) and the lowest CRPS and RMSE in the Kumpula-to-Hyyti\"al\"a direction (444.4\,cm$^{-3}$ and 1188.6\,cm$^{-3}$, respectively). In the Hyyti\"al\"a-to-Kumpula direction, the Drifting Model has the lowest CRPS (1174.9\,cm$^{-3}$) and FM+CFG the lowest RMSE (2494.4\,cm$^{-3}$). Energy Matching occasionally produces extremely large samples under transfer. Its RMSE, which squares these values, is therefore not reported in either direction, and in the Kumpula-to-Hyyti\"al\"a direction its CRPS is also numerically unreliable and is omitted.

The physical constraints behave as in the local tests (Table~\ref{tab:physics_crossdomain}). From Hyyti\"al\"a to Kumpula, GDE lowers the CRPS of FM from 1204.7\,cm$^{-3}$ to 1188.9\,cm$^{-3}$, and number+GDE gives the lowest value (1174.0\,cm$^{-3}$), at a lower correlation than GDE alone (0.723 against 0.732). In the reverse direction, GDE raises the correlation from 0.660 to 0.667 and lowers the CRPS from 830.8\,cm$^{-3}$ to 732.9\,cm$^{-3}$; mass+GDE gives a similar 732.4\,cm$^{-3}$. These improvements are modest and do not remove the component discrepancies described in Appendix~\ref{sec:modal}.

\begin{figure*}[t]
\includegraphics[width=0.9\textwidth]{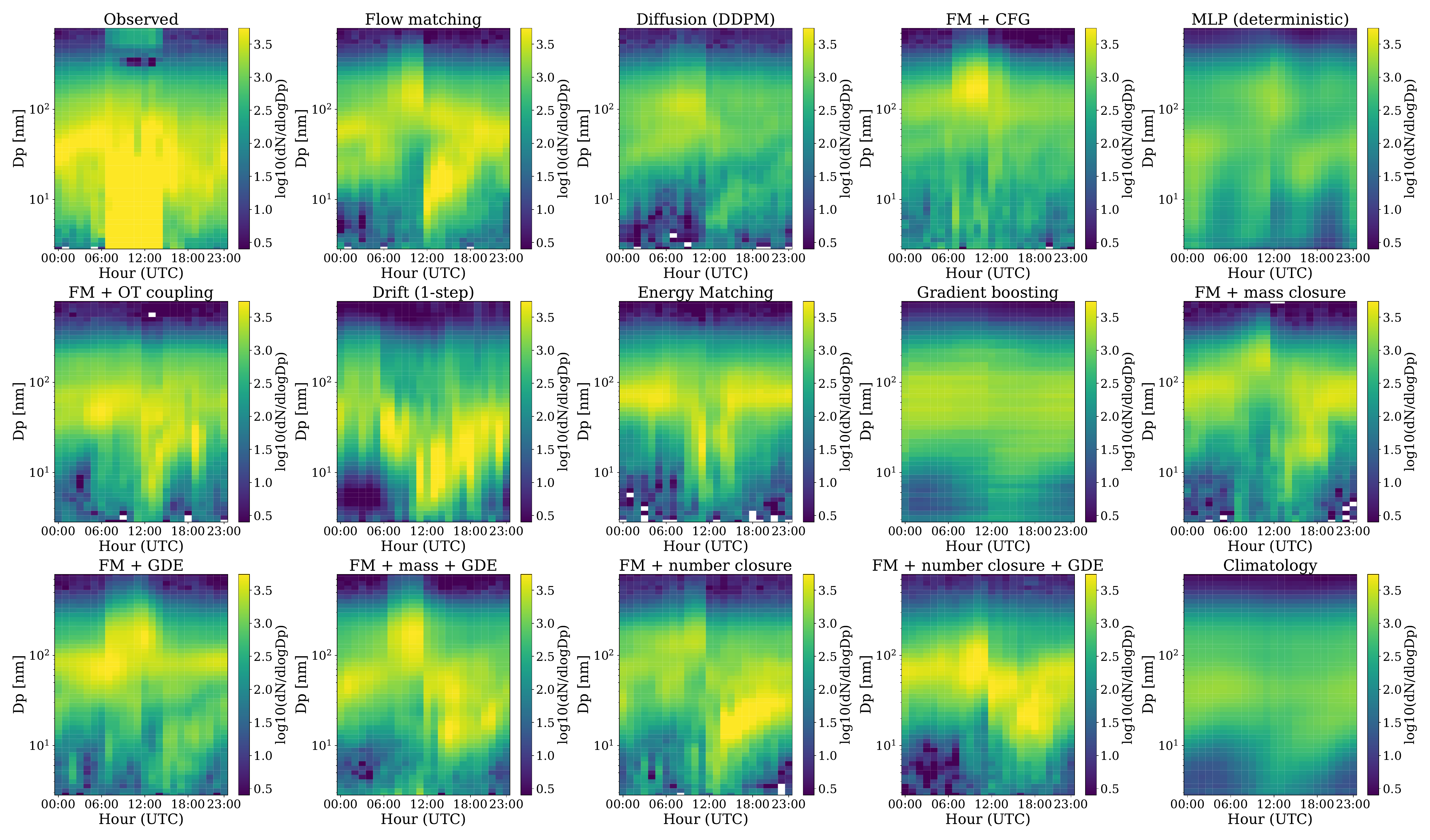}
\caption{\textbf{No Hyyti\"al\"a-trained model reproduces the daytime UFP enhancement at Kumpula, most likely caused by traffic emissions.} Zero-shot application to SMEAR~III for one day, compared with the observation (top left). Each panel is one 24\,h window generated as a whole from the inputs of that window, not a sequence of separate hourly estimates; features that appear at the same hour in several models, such as the change near 12\,UTC, therefore reflect the common inputs of the window or the diurnal pattern learned at the training site. Axes and color scale are as in Fig.~\ref{fig:grid_extrap}.}
\label{fig:grid_ood}
\end{figure*}

\begin{table*}[t]
\caption{Cross-site (zero-shot) full-PNSD scores over all valid target-site windows. Correlation uses log-transformed PNSDs; CRPS and RMSE are in cm$^{-3}$. $\dagger$ RMSE is not reported for Energy Matching in either direction because it is dominated by rare, extremely large samples. $\ddagger$ In the Kumpula-to-Hyyti\"al\"a direction, Energy Matching's CRPS is also omitted (see text). Best values are in bold.}
\label{tab:crossdomain}
\centering
\begin{minipage}[t]{0.49\textwidth}
\centering
\textbf{(a) Hyyti\"al\"a-trained $\to$ Kumpula}\par
\scriptsize
\begin{tabular}{lrrr}
\tophline
Model & corr. $\uparrow$ & CRPS $\downarrow$ & RMSE $\downarrow$ \\
\middlehline
Climatology & 0.714 & 1507.2 & 2705.8 \\
Gradient Boosting & 0.701 & 1387.1 & 2602.9 \\
MLP & 0.658 & 1399.4 & 2689.4 \\
Diffusion Model & \textbf{0.792} & 1228.5 & 2547.3 \\
Drifting Model & 0.748 & \textbf{1174.9} & 2515.3 \\
Energy Matching & 0.756 & 1221.3 & --$^\dagger$ \\
Flow Matching & 0.724 & 1204.7 & 2503.5 \\
FM + OT & 0.713 & 1253.0 & 2542.6 \\
FM + CFG & 0.723 & 1193.5 & \textbf{2494.4} \\
\bottomhline
\end{tabular}
\end{minipage}\hfill
\begin{minipage}[t]{0.49\textwidth}
\centering
\textbf{(b) Kumpula-trained $\to$ Hyyti\"al\"a}\par
\scriptsize
\begin{tabular}{lrrr}
\tophline
Model & corr. $\uparrow$ & CRPS $\downarrow$ & RMSE $\downarrow$ \\
\middlehline
Climatology & 0.670 & 795.3 & 1222.1 \\
Gradient Boosting & 0.679 & 999.2 & 1505.6 \\
MLP & 0.655 & 1267.7 & 2486.4 \\
Diffusion Model & \textbf{0.689} & \textbf{444.4} & \textbf{1188.6} \\
Drifting Model & 0.649 & 697.8 & 2108.6 \\
Energy Matching & 0.662 & --$^\ddagger$ & --$^\dagger$ \\
Flow Matching & 0.660 & 830.8 & 1945.4 \\
FM + OT & 0.667 & 820.9 & 1830.3 \\
FM + CFG & 0.660 & 757.7 & 1843.8 \\
\bottomhline
\end{tabular}
\end{minipage}
\end{table*}

\begin{table*}[t]
\caption{Physical-constraint ablation under zero-shot transfer, over all valid target-site windows. Columns follow Table~\ref{tab:physics_indomain}, omitting PNSD RMSE; CRPS is in cm$^{-3}$ and mass RMSE in $\mu$g\,m$^{-3}$. These periods differ from the local temporal holdouts and the paired tests. Best values are in bold.}
\label{tab:physics_crossdomain}
\centering
\begin{minipage}[t]{0.49\textwidth}
\centering
\textbf{(a) Hyyti\"al\"a-trained $\to$ Kumpula}\par
\scriptsize
\begin{tabular}{lrrr}
\tophline
Model & corr. & CRPS & mass RMSE \\
\middlehline
Flow Matching & 0.724 & 1204.7 & 1.29 \\
FM + mass & 0.696 & 1255.8 & 1.32 \\
FM + GDE & \textbf{0.732} & 1188.9 & \textbf{1.25} \\
FM + mass + GDE & 0.728 & 1185.9 & \textbf{1.25} \\
FM + number & 0.702 & 1227.1 & 1.26 \\
FM + number + GDE & 0.723 & \textbf{1174.0} & \textbf{1.25} \\
\bottomhline
\end{tabular}
\end{minipage}\hfill
\begin{minipage}[t]{0.49\textwidth}
\centering
\textbf{(b) Kumpula-trained $\to$ Hyyti\"al\"a}\par
\scriptsize
\begin{tabular}{lrrr}
\tophline
Model & corr. & CRPS & mass RMSE \\
\middlehline
Flow Matching & 0.660 & 830.8 & 1.78 \\
FM + mass & 0.664 & 927.3 & 2.10 \\
FM + GDE & 0.667 & 732.9 & 1.56 \\
FM + mass + GDE & 0.668 & \textbf{732.4} & \textbf{1.03} \\
FM + number & 0.665 & 913.6 & 2.32 \\
FM + number + GDE & \textbf{0.672} & 817.1 & 1.29 \\
\bottomhline
\end{tabular}
\end{minipage}
\end{table*}

\section{Derived quantities and episode verification}
\label{app:quantities}
\begin{figure*}[t]
\centering
\includegraphics[width=0.96\textwidth]{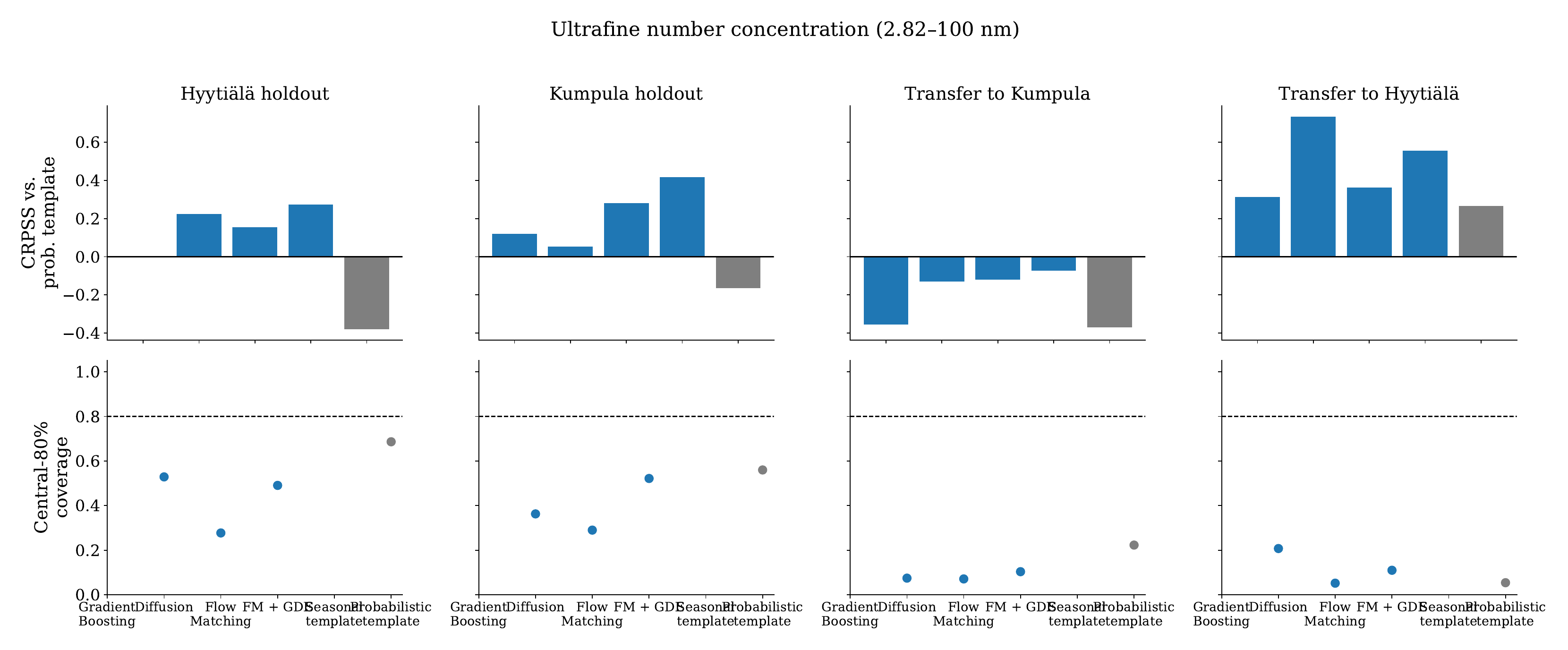}\\[0.5em]
\includegraphics[width=0.96\textwidth]{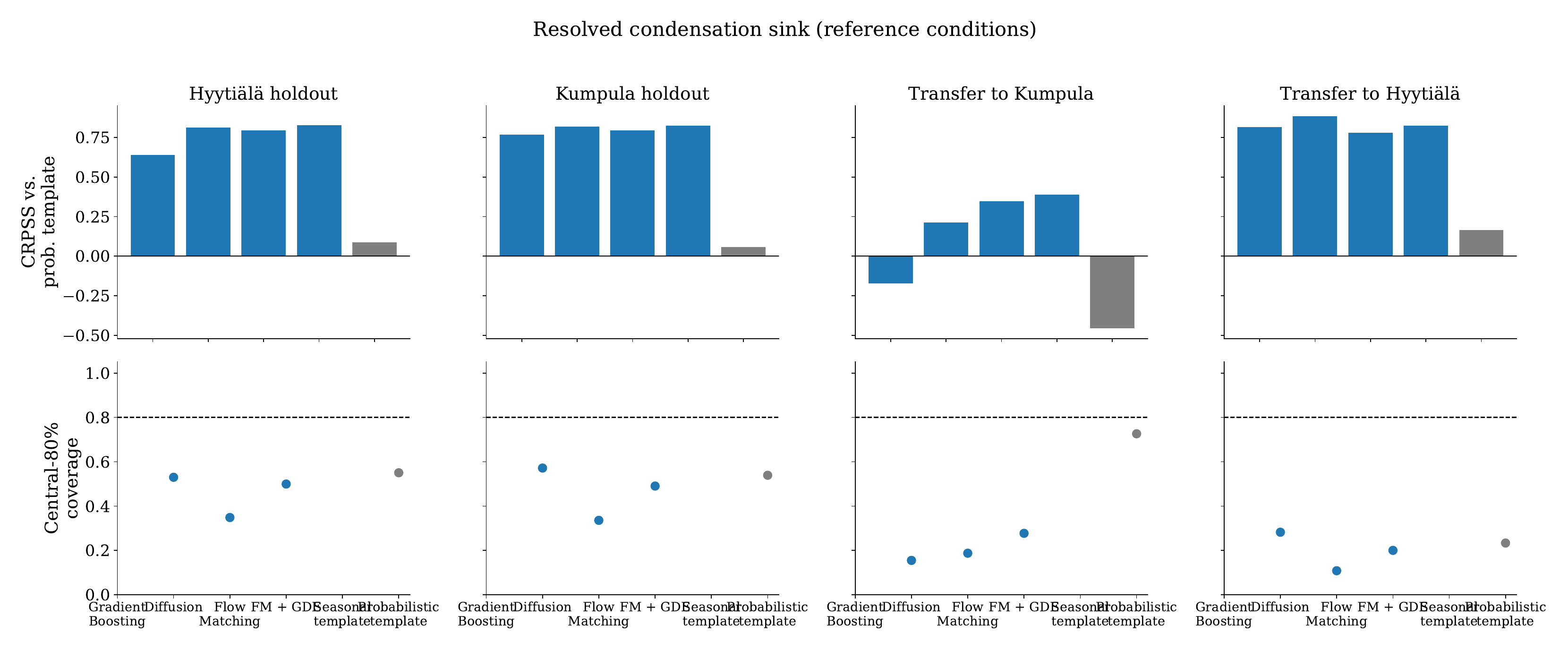}
\caption{\textbf{Local skill is positive for both quantities, but boreal-to-urban transfer loses all skill for \ncufp.} Derived-quantity skill and calibration on paired target-site holdouts: \ncufp{} (upper group) and the resolved CS (lower group). Each group shows CRPSS relative to the source-training historical ensemble above central-80\,\% coverage, with methods on the vertical axis. Dashed lines mark zero skill and nominal coverage of 0.80. Values are from Table~\ref{tab:atmquant}.}
\label{fig:atmquant}
\end{figure*}

The derived quantities of Sect.~\ref{sec:res_atmquant} are computed separately for every ensemble member before scores are calculated, so the predicted distribution of each quantity retains the dependence between bins. \ncufp{} covers 2.82--100\,nm, and CS uses Eq.~(\ref{eq:cs}) at the reference conditions of Appendix~\ref{sec:sinks}. Table~\ref{tab:atmquant} gives the full scores for the four evaluation settings: the two local holdouts and the two transfer directions, all on the same target-site windows. The seasonal template and the historical ensemble are always built from the training data of the source station. A positive CRPSS therefore means an improvement over what the source archive alone would provide, not over a locally trained model; Fig.~\ref{fig:atmquant} summarizes skill and coverage.

\begin{table*}[t]
\caption{Derived-quantity scores on paired target-site temporal holdouts: observed-range \ncufp{} and the resolved CS. CRPSS is relative to the source-training historical ensemble (``probabilistic template''); bold marks the best CRPSS among the four learned models in each block. Deterministic predictions have no interval coverage or width. Unlike Tables~\ref{tab:crossdomain} and \ref{tab:physics_crossdomain}, transfers here use the same target windows as the local models.}
\label{tab:atmquant}
\centering
\setlength{\tabcolsep}{3pt}
\begin{minipage}[t]{0.49\textwidth}
\centering
\textbf{(a) Hyyti\"al\"a holdout}\par
\scriptsize
\begin{tabular}{lrrrr}
\tophline
Model & CRPS $\downarrow$ & Cov80\% & Width80\% & CRPSS $\uparrow$ \\
\middlehline
\multicolumn{5}{l}{\textit{\ncufp{} (2.82--100\,nm), cm$^{-3}$}} \\
Gradient Boosting & 250 & -- & -- & +0.00 \\
Diffusion Model & 194 & 0.53 & 521 & +0.22 \\
Flow Matching & 211 & 0.28 & 249 & +0.15 \\
FM + GDE & 181 & 0.49 & 391 & \textbf{+0.27} \\
Seasonal template & 345 & -- & -- & -0.38 \\
Probabilistic template & 250 & 0.69 & 1158 & -- \\
\multicolumn{5}{l}{\textit{CS, s$^{-1}$}} \\
Gradient Boosting & $1.73\times10^{-4}$ & -- & -- & +0.64 \\
Diffusion Model & $0.90\times10^{-4}$ & 0.53 & $2.06\times10^{-4}$ & +0.81 \\
Flow Matching & $0.98\times10^{-4}$ & 0.35 & $1.25\times10^{-4}$ & +0.80 \\
FM + GDE & $0.83\times10^{-4}$ & 0.50 & $1.76\times10^{-4}$ & \textbf{+0.83} \\
Seasonal template & $4.39\times10^{-4}$ & -- & -- & +0.09 \\
Probabilistic template & $4.80\times10^{-4}$ & 0.55 & $2.53\times10^{-3}$ & -- \\
\bottomhline
\end{tabular}
\end{minipage}\hfill
\begin{minipage}[t]{0.49\textwidth}
\centering
\textbf{(b) Kumpula holdout}\par
\scriptsize
\begin{tabular}{lrrrr}
\tophline
Model & CRPS $\downarrow$ & Cov80\% & Width80\% & CRPSS $\uparrow$ \\
\middlehline
\multicolumn{5}{l}{\textit{\ncufp{} (2.82--100\,nm), cm$^{-3}$}} \\
Gradient Boosting & 904 & -- & -- & +0.12 \\
Diffusion Model & 973 & 0.36 & 1110 & +0.05 \\
Flow Matching & 739 & 0.29 & 1088 & +0.28 \\
FM + GDE & 599 & 0.52 & 1583 & \textbf{+0.42} \\
Seasonal template & 1198 & -- & -- & -0.17 \\
Probabilistic template & 1028 & 0.56 & 4070 & -- \\
\multicolumn{5}{l}{\textit{CS, s$^{-1}$}} \\
Gradient Boosting & $1.97\times10^{-4}$ & -- & -- & +0.77 \\
Diffusion Model & $1.54\times10^{-4}$ & 0.57 & $3.56\times10^{-4}$ & +0.82 \\
Flow Matching & $1.74\times10^{-4}$ & 0.34 & $2.73\times10^{-4}$ & +0.79 \\
FM + GDE & $1.49\times10^{-4}$ & 0.49 & $4.06\times10^{-4}$ & \textbf{+0.82} \\
Seasonal template & $7.95\times10^{-4}$ & -- & -- & +0.06 \\
Probabilistic template & $8.44\times10^{-4}$ & 0.54 & $4.03\times10^{-3}$ & -- \\
\bottomhline
\end{tabular}
\end{minipage}

\begin{minipage}[t]{0.49\textwidth}
\centering
\textbf{(c) Hyyti\"al\"a-trained $\to$ Kumpula}\par
\scriptsize
\begin{tabular}{lrrrr}
\tophline
Model & CRPS $\downarrow$ & Cov80\% & Width80\% & CRPSS $\uparrow$ \\
\middlehline
\multicolumn{5}{l}{\textit{\ncufp{} (2.82--100\,nm), cm$^{-3}$}} \\
Gradient Boosting & 2294 & -- & -- & -0.36 \\
Diffusion Model & 1914 & 0.07 & 596 & -0.13 \\
Flow Matching & 1896 & 0.07 & 324 & -0.12 \\
FM + GDE & 1815 & 0.10 & 500 & \textbf{-0.07} \\
Seasonal template & 2321 & -- & -- & -0.37 \\
Probabilistic template & 1693 & 0.22 & 1158 & -- \\
\multicolumn{5}{l}{\textit{CS, s$^{-1}$}} \\
Gradient Boosting & $6.26\times10^{-4}$ & -- & -- & -0.17 \\
Diffusion Model & $4.20\times10^{-4}$ & 0.16 & $3.34\times10^{-4}$ & +0.21 \\
Flow Matching & $3.49\times10^{-4}$ & 0.19 & $2.03\times10^{-4}$ & +0.35 \\
FM + GDE & $3.26\times10^{-4}$ & 0.28 & $2.94\times10^{-4}$ & \textbf{+0.39} \\
Seasonal template & $7.76\times10^{-4}$ & -- & -- & -0.45 \\
Probabilistic template & $5.33\times10^{-4}$ & 0.73 & $2.53\times10^{-3}$ & -- \\
\bottomhline
\end{tabular}
\end{minipage}\hfill
\begin{minipage}[t]{0.49\textwidth}
\centering
\textbf{(d) Kumpula-trained $\to$ Hyyti\"al\"a}\par
\scriptsize
\begin{tabular}{lrrrr}
\tophline
Model & CRPS $\downarrow$ & Cov80\% & Width80\% & CRPSS $\uparrow$ \\
\middlehline
\multicolumn{5}{l}{\textit{\ncufp{} (2.82--100\,nm), cm$^{-3}$}} \\
Gradient Boosting & 1378 & -- & -- & +0.31 \\
Diffusion Model & 532 & 0.21 & 806 & \textbf{+0.74} \\
Flow Matching & 1281 & 0.05 & 686 & +0.36 \\
FM + GDE & 891 & 0.11 & 1053 & +0.56 \\
Seasonal template & 1474 & -- & -- & +0.27 \\
Probabilistic template & 2010 & 0.05 & 4070 & -- \\
\multicolumn{5}{l}{\textit{CS, s$^{-1}$}} \\
Gradient Boosting & $2.08\times10^{-4}$ & -- & -- & +0.82 \\
Diffusion Model & $1.33\times10^{-4}$ & 0.28 & $1.97\times10^{-4}$ & \textbf{+0.88} \\
Flow Matching & $2.49\times10^{-4}$ & 0.11 & $1.59\times10^{-4}$ & +0.78 \\
FM + GDE & $1.97\times10^{-4}$ & 0.20 & $2.36\times10^{-4}$ & +0.83 \\
Seasonal template & $9.45\times10^{-4}$ & -- & -- & +0.17 \\
Probabilistic template & $1.13\times10^{-3}$ & 0.23 & $4.03\times10^{-3}$ & -- \\
\bottomhline
\end{tabular}
\end{minipage}
\end{table*}

\subsection{Sustained high-number episodes}
\label{sec:res_episode}
An ensemble can also provide probabilities of events. As an example, we define a sustained high-number episode for a 24\,h sequence by its largest 3\,h running mean of \ncufp, $Q$, and call the window an event if $Q$ exceeds a threshold $q$. The threshold is the 90th percentile of $Q$ in the training data of the source station and is kept unchanged under transfer. The predicted event probability is the fraction of ensemble members whose $Q$ exceeds $q$, and its accuracy is measured by the Brier score, the mean squared difference between predicted probability and observed outcome \citep{brier1950}. To avoid counting the same episode in many overlapping windows, we use one window per day, giving 60 daily cases per setting; consecutive days can still be correlated, and the one-day bootstrap blocks used for the intervals do not account for longer dependence. The event is a statistical definition, not a classification of new particle formation and not a health threshold.

\begin{figure*}[htbp]
\includegraphics[width=0.95\textwidth]{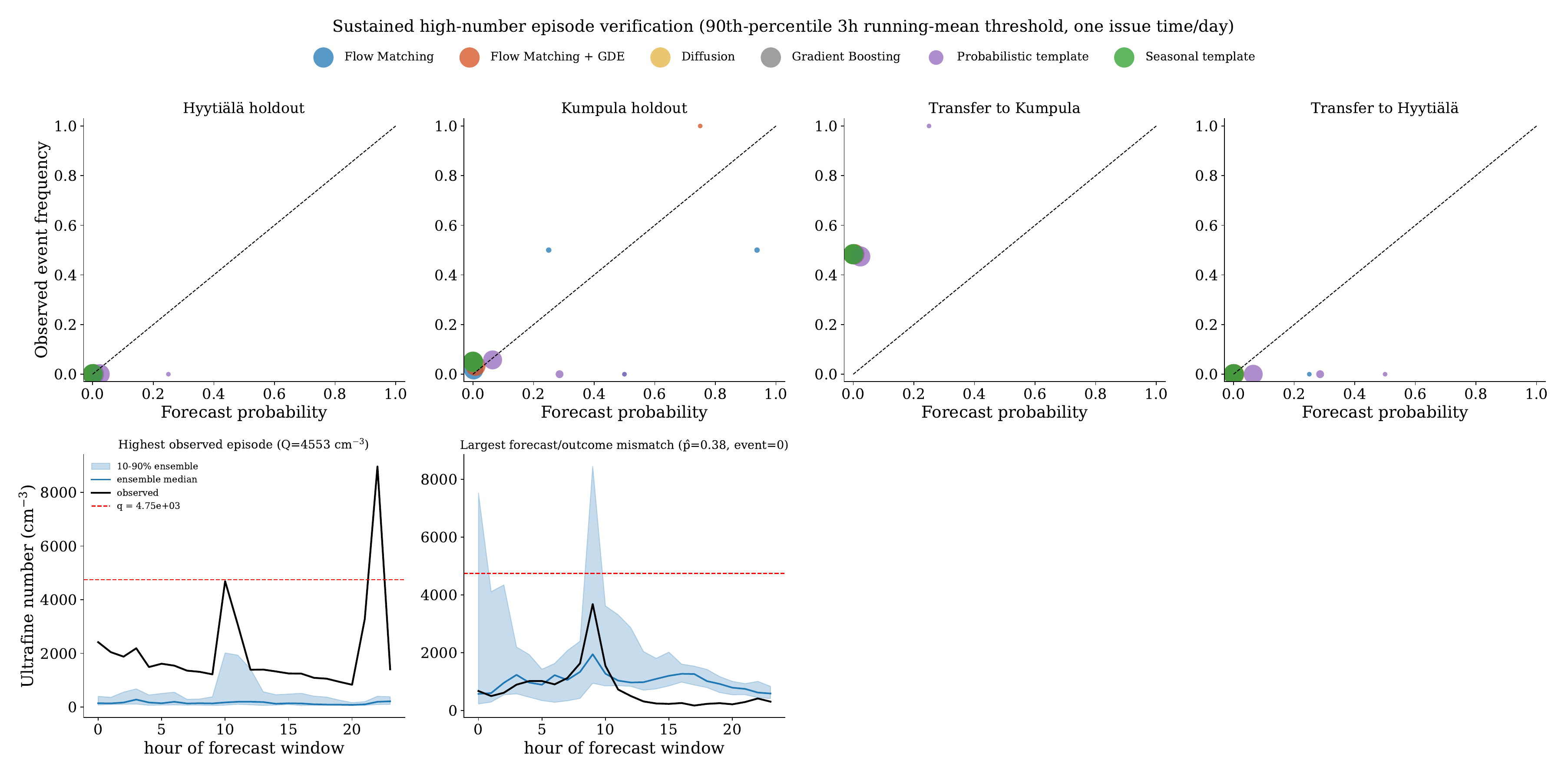}
\caption{\textbf{The 60-day holdouts contain too few high-number episodes for reliable probability verification.} Top row: reliability diagram per evaluation setting (mean predicted probability against observed event frequency; marker area scales with the number of days in each bin, and sparse bins are shown as they are). Bottom row: two 24\,h examples for the Hyyti\"al\"a FM+GDE model, the highest observed episode (left) and the day with the largest mismatch between prediction and outcome, a failure case (right). The shaded band is the 10--90\,\% ensemble interval and the dashed line the threshold $q$.}
\label{fig:episode}
\end{figure*}

The holdout periods contain few events. At Hyyti\"al\"a (threshold 4745\,cm$^{-3}$) no day qualifies, and at Kumpula (threshold 13\,639\,cm$^{-3}$) only 3 of 60 days do. Under transfer, the Hyyti\"al\"a threshold is exceeded on 29 of 60 days at Kumpula, whereas the Kumpula threshold is never exceeded at Hyyti\"al\"a, so a percentile threshold does not keep its rarity in a new environment. Near-zero Brier scores for samples without events do not show that events can be detected. Kumpula's local scores (0.031--0.067) rest on three events and cannot support a ranking, and in Hyyti\"al\"a-to-Kumpula transfer, scores of 0.454--0.484 are hardly better than always predicting no event (Brier score $29/60=0.483$). Table~\ref{tab:episode} and Fig.~\ref{fig:episode} document the available sample; reliable verification of event probabilities requires longer holdout periods.

\begin{table*}[htbp]
\caption{Sustained high-number episode verification, with evaluation settings in rows and methods in columns. Cells give the Brier score above its 95\,\% block-bootstrap interval (block length 1 day). Event counts give the observed days exceeding the source threshold $q$ (cm$^{-3}$).}
\label{tab:episode}
\centering\small
\renewcommand{\arraystretch}{1.2}
\setlength{\tabcolsep}{3pt}
\begin{tabular*}{\textwidth}{@{\extracolsep{\fill}}lcccccc@{}}
\tophline
Setting & \shortstack{Gradient\\boosting} & Diffusion & FM & FM + GDE &
\shortstack{Seasonal\\template} & \shortstack{Historical\\ensemble} \\
\middlehline
\shortstack[l]{Hyyti\"al\"a local\\$q=4745$; 0/60 events} & \shortstack{0.000\\{}[0.000,0.000]} & \shortstack{0.000\\{}[0.000,0.001]} & \shortstack{0.001\\{}[0.000,0.002]} & \shortstack{0.000\\{}[0.000,0.001]} & \shortstack{0.000\\{}[0.000,0.000]} & \shortstack{0.004\\{}[0.002,0.007]} \\
\middlehline
\shortstack[l]{Kumpula local\\$q=13639$; 3/60 events} & \shortstack{0.050\\{}[0.000,0.117]} & \shortstack{0.050\\{}[0.000,0.117]} & \shortstack{0.044\\{}[0.006,0.094]} & \shortstack{0.031\\{}[0.001,0.077]} & \shortstack{0.050\\{}[0.000,0.117]} & \shortstack{0.067\\{}[0.023,0.124]} \\
\middlehline
\shortstack[l]{Hyyti\"al\"a $\to$ Kumpula\\$q=4745$; 29/60 events} & \shortstack{0.483\\{}[0.367,0.617]} & \shortstack{0.483\\{}[0.367,0.617]} & \shortstack{0.484\\{}[0.367,0.617]} & \shortstack{0.484\\{}[0.367,0.617]} & \shortstack{0.483\\{}[0.367,0.617]} & \shortstack{0.454\\{}[0.344,0.576]} \\
\middlehline
\shortstack[l]{Kumpula $\to$ Hyyti\"al\"a\\$q=13639$; 0/60 events} & \shortstack{0.000\\{}[0.000,0.000]} & \shortstack{0.000\\{}[0.000,0.001]} & \shortstack{0.001\\{}[0.000,0.003]} & \shortstack{0.000\\{}[0.000,0.000]} & \shortstack{0.000\\{}[0.000,0.000]} & \shortstack{0.021\\{}[0.012,0.032]} \\
\bottomhline
\end{tabular*}
\end{table*}

\section{Fitted size components}
\label{sec:modal}\label{sec:res_modal}\label{sec:res_modal_hyy}
\begin{figure*}[htbp]
\includegraphics[width=0.9\textwidth]{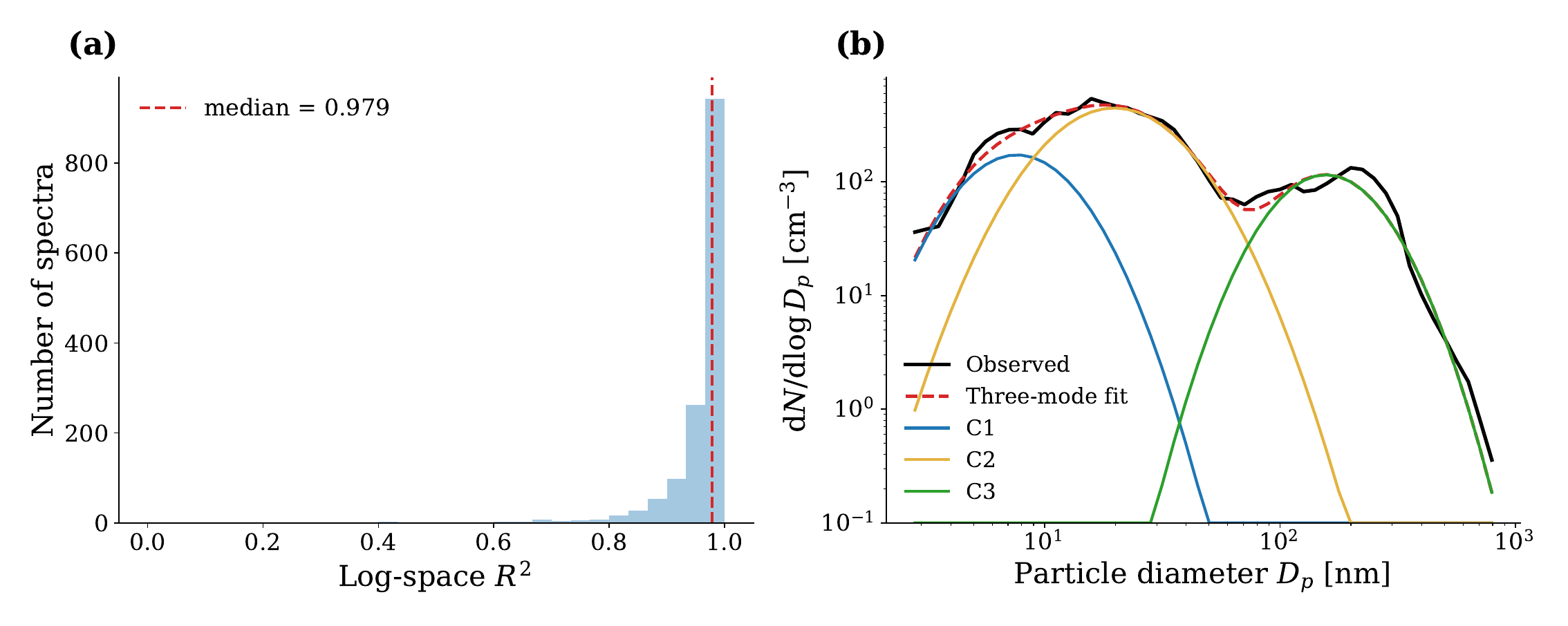}
\caption{\textbf{Three log-normal components describe the Hyyti\"al\"a holdout PNSDs well.} Left: distribution of log-space fit $R^2$, with the median marked. Right: an observed PNSD and its ordered fitted components C1--C3 on the 2.82--794\,nm grid.}
\label{fig:mode_fit_hyy}
\end{figure*}

To see which part of the PNSD is lost under transfer, we describe each observed and predicted PNSD by three log-normal components \citep{hussein2005},
\begin{equation}
n(D_p)=\sum_{k=1}^{3}\frac{N_k}{\sqrt{2\pi}\,\log_{10}\sigma_k}
\exp\!\left[-\frac{\left(\log_{10}D_p-\log_{10}D_{g,k}\right)^2}{2\left(\log_{10}\sigma_k\right)^2}\right],
\end{equation}
where $N_k$ is the number concentration, $D_{g,k}$ the geometric mean diameter and $\sigma_k$ the geometric standard deviation of component $k$. The parameters are fitted by minimizing relative residuals, so that the smallest particles carry weight comparable to the peak of the PNSD, and the ranges allowed for $D_{g,k}$ do not overlap, so that each component keeps its identity across PNSDs. We label the components C1--C3 in order of increasing diameter. They are fitted components, not identified sources or the conventional Aitken, accumulation and coarse modes, and the 2.82--794\,nm grid does not resolve a coarse mode. The fits describe the PNSDs well, with median log-space $R^2$ values of 0.978--0.991 (Fig.~\ref{fig:mode_fit_hyy}), although a good fit does not guarantee that the component parameters are unique.

For each component, the component number $N_k$ is the integral of the fitted log-normal, including its tails beyond the observed grid. We summarize the amplitude bias of a model by $R_k$, the median over windows of the ratio of predicted to observed $N_k$, excluding windows with observed $N_k$ below 1\,cm$^{-3}$. The transfer statistic
\begin{equation}
T_k=\frac{R_k^{\rm cross}}{R_k^{\rm same}}
\end{equation}
divides the ratio at the other station by the ratio at the training station; $T_k=1$ means that transfer leaves the relative amplitude bias unchanged, not that the transferred prediction is unbiased.

\begin{figure*}[t]
\includegraphics[width=0.9\textwidth]{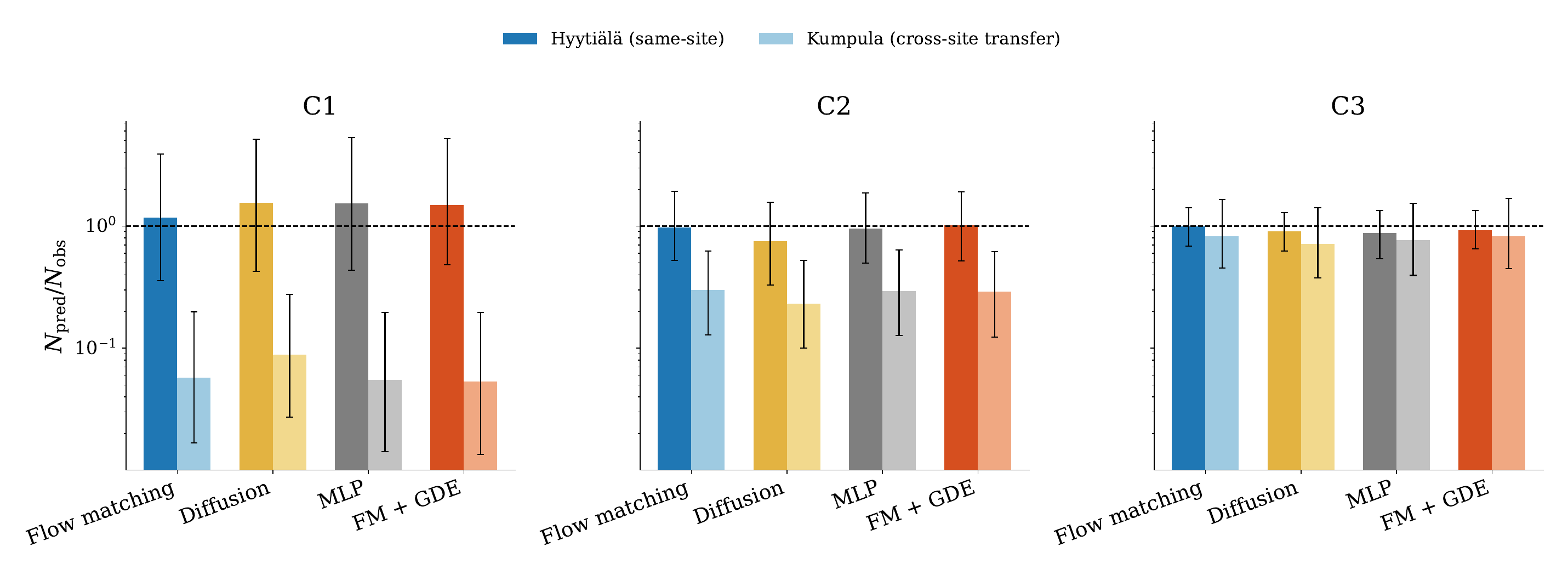}
\caption{\textbf{Boreal-to-urban transfer changes the smallest fitted component most.} Component-number ratios for Hyyti\"al\"a-trained models, evaluated at Hyyti\"al\"a (solid) and Kumpula (translucent). Bars show the median predicted/observed number and whiskers the interquartile range. C1--C3 are ordered fitted components on the 2.82--794\,nm grid; these raw ratios differ from the normalized ratios in Table~\ref{tab:modes_transfer}.}
\label{fig:transfer}
\end{figure*}

For Hyyti\"al\"a-trained FM, the local ratios $R_k$ are 1.18, 0.98 and 0.99 for C1--C3, and the ratios at Kumpula are 0.058, 0.301 and 0.829. For Kumpula-trained FM, the local ratios are 1.07, 1.30 and 0.98, and the ratios at Hyyti\"al\"a are 17.20, 3.17 and 1.15. Across all FM variants, the normalized C1 ratios span 0.035--0.054 in the boreal-to-urban direction and 13.8--16.1 in the reverse direction, whereas C3 stays near unity and C2 lies in between (Table~\ref{tab:modes_transfer}, Fig.~\ref{fig:transfer}). Transfer therefore affects the smallest particles most, which is consistent with the weak constraint that bulk mass places on small-particle number; the fits do not attribute these differences to particular sources. Locally, the median component ratios of FM are close to one (Table~\ref{tab:modes_hyy}), but the interquartile range for C1 is broad (0.36--3.89), so agreement in the median does not imply accurate components in individual windows.

\begin{table*}[t]
\caption{Normalized component transfer ratios $T_k$: cross-site predicted/observed component number divided by the same model's ratio at its training site. C1--C3 are ordered fitted components. Unity denotes an unchanged relative bias, not necessarily an unbiased target-site prediction.}
\label{tab:modes_transfer}
\centering
\begin{minipage}[t]{0.49\textwidth}
\centering
\textbf{(a) Hyyti\"al\"a-trained $\to$ Kumpula}\par
\scriptsize
\begin{tabular}{lccc}
\tophline
Model & C1 & C2 & C3 \\
\middlehline
Flow Matching & 0.049 & 0.309 & 0.837 \\
FM + mass & 0.035 & 0.285 & 0.859 \\
FM + GDE & 0.036 & 0.284 & 0.888 \\
FM + mass + GDE & 0.049 & 0.313 & 0.882 \\
FM + number & 0.047 & 0.272 & \textbf{0.946} \\
FM + number + GDE & 0.054 & 0.289 & 0.924 \\
\bottomhline
\end{tabular}
\end{minipage}\hfill
\begin{minipage}[t]{0.49\textwidth}
\centering
\textbf{(b) Kumpula-trained $\to$ Hyyti\"al\"a}\par
\scriptsize
\begin{tabular}{lccc}
\tophline
Model & C1 & C2 & C3 \\
\middlehline
Flow Matching & 16.1 & 2.44 & 1.18 \\
FM + mass & 15.3 & 2.54 & 1.23 \\
FM + GDE & 13.8 & 2.55 & 1.03 \\
FM + mass + GDE & 15.5 & 2.49 & 1.10 \\
FM + number & 14.5 & 2.72 & 1.22 \\
FM + number + GDE & 15.1 & 2.56 & \textbf{1.00} \\
\bottomhline
\end{tabular}
\end{minipage}
\end{table*}

\begin{table*}[t]
\caption{Median ratio of predicted to observed component number at Hyyti\"al\"a (temporal holdout), with interquartile range. Unity indicates an unbiased median component amplitude; it does not imply accurate individual PNSDs.}
\label{tab:modes_hyy}
\centering
\begin{tabular}{lccc}
\tophline
Model & C1 & C2 & C3 \\
\middlehline
Flow Matching & 1.18 (0.36--3.89) & 0.98 (0.53--1.93) & 0.99 (0.68--1.42) \\
Diffusion Model & 1.54 (0.43--5.12) & 0.75 (0.33--1.57) & 0.91 (0.62--1.29) \\
FM + CFG & 1.26 (0.39--4.44) & 0.84 (0.45--1.64) & 1.00 (0.69--1.43) \\
MLP & 1.54 (0.43--5.32) & 0.95 (0.50--1.88) & 0.88 (0.54--1.34) \\
FM + GDE & 1.49 (0.48--5.21) & 1.02 (0.52--1.90) & 0.93 (0.65--1.34) \\
\bottomhline
\end{tabular}
\end{table*}

\section{SOSAA-FP and its comparison diagnostic}
\label{app:sosaa_seasonal}
SOSAA is a one-dimensional column model that simulates the evolution of gases and aerosol particles in the atmospheric boundary layer by combining meteorology, emissions, gas-phase chemistry and aerosol dynamics \citep{boy2011,clusius2026}. In the SOSAA-FP configuration of \citet{clusius2026}, the column is simulated along backward air-mass trajectories arriving at Hyyti\"al\"a, so that the simulated PNSD reflects the emissions and processes encountered upstream of the station. We use the archived simulations of \citet{clusius2026}, whose 60-bin grid (1.07--2000\,nm) contains the 50 diameters used here, so that observed, simulated and generated PNSDs can be compared bin by bin without interpolation. The matched windows of Sect.~\ref{sec:sosaa} are those of the interpolation split (Appendix~\ref{app:data}) that fall within the simulated period.

\begin{figure*}[htbp]
\includegraphics[width=\textwidth]{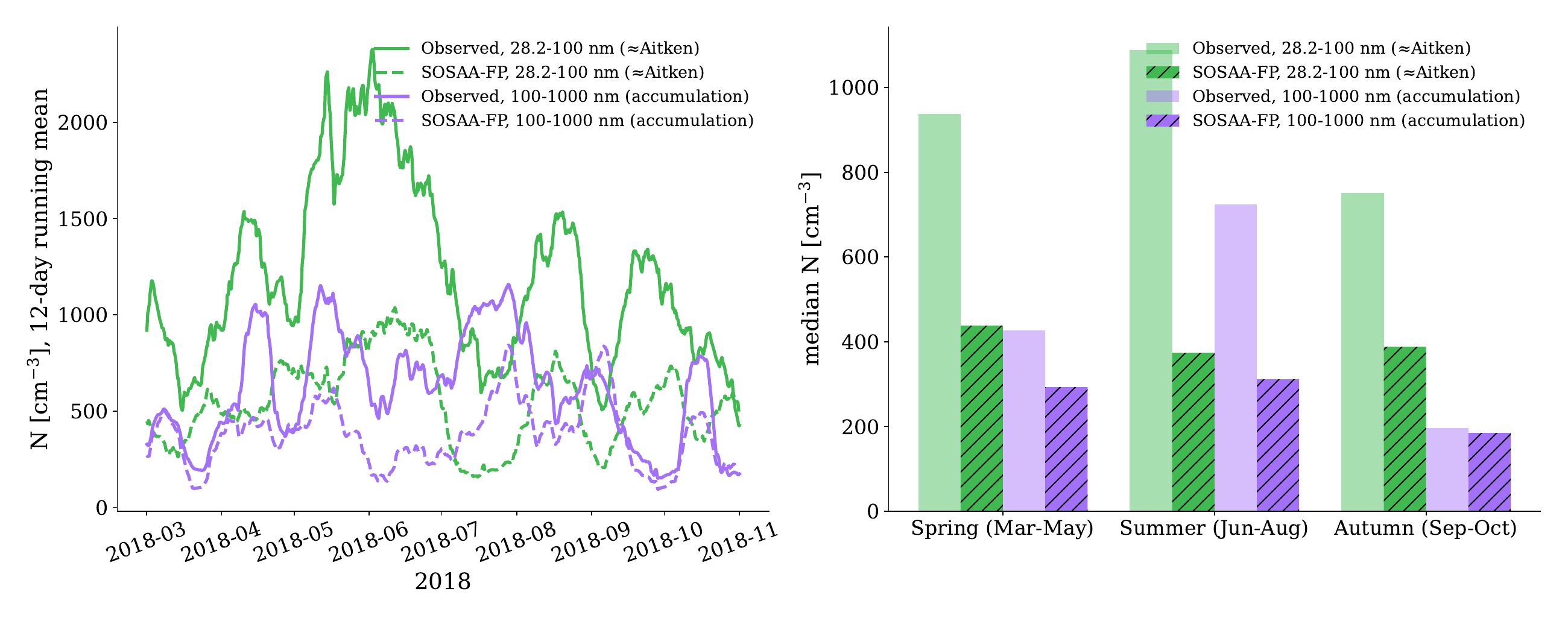}
\caption{\textbf{The validation diagnostic of \citet{clusius2026} applied to the matched Hyyti\"al\"a/SOSAA-FP period.} Twelve-day running means (left) and seasonal medians (right) for the nominal 28.2--100 and 100--1000\,nm ranges (the latter truncated at 794\,nm). Solid lines and bars: observations; dashed lines and hatched bars: SOSAA-FP.}
\label{fig:sosaa_running_mean}
\end{figure*}

Hourly PNSDs emphasize short-term differences. To check whether the disagreement in Sect.~\ref{sec:sosaa} persists at longer time scales, we also apply the validation diagnostic of \citet{clusius2026}: 12\,d running means and seasonal medians of the number concentration in two size ranges (Fig.~\ref{fig:sosaa_running_mean}). We use nominal ranges of 28.2--100 and 100--1000\,nm, chosen to match the 30--100 and 100--1000\,nm ranges of that study as closely as our bin edges allow. Both the observed and the simulated series are computed on the 50-bin analysis grid, so the upper range is truncated at 794\,nm in both.

\section{Sampling cost}
\label{sec:appendix_nfe}
\begin{figure}[htbp]
\includegraphics[width=0.9\textwidth]{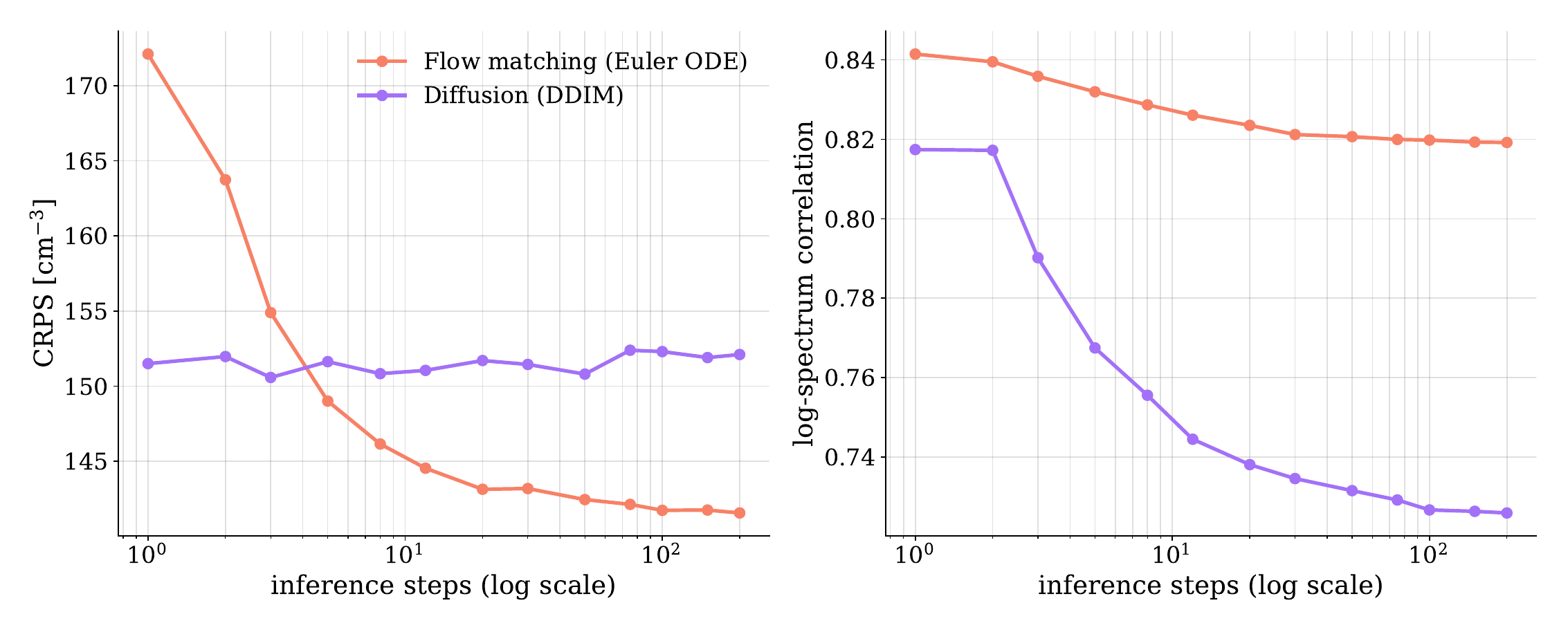}
\caption{\textbf{Flow Matching needs about 50 integration steps, whereas diffusion reaches its CRPS plateau within a few steps at a cost in correlation.} (a)~CRPS and (b)~correlation of log-transformed PNSDs on the Hyyti\"al\"a temporal holdout as a function of the number of function evaluations (Euler steps for Flow Matching, DDIM denoising steps for diffusion).}
\label{fig:nfe}
\end{figure}

Iterative generative models trade accuracy against cost through the number of network evaluations (NFE) per sample. For FM, the NFE is the number of Euler steps $K$ in Eq.~(\ref{eq:fm_sampling}); for diffusion, it is the number of denoising steps, which we vary using the deterministic DDIM sampler; the Drifting Model always uses a single evaluation. Figure~\ref{fig:nfe} shows that FM needs about 50 steps for its CRPS to converge, whereas its correlation hardly depends on the step count. Diffusion reaches its CRPS plateau within a few steps, but its correlation decreases as the number of steps grows. The two methods therefore trade point accuracy against probabilistic skill along the sampling budget, and neither dominates. The Drifting Model is not shown because its cost is fixed.

\codedataavailability{Repository DOI (Zenodo) for the model and analysis code (\url{10.5281/zenodo.23540799}) SMEAR data are available via the SmartSMEAR API (\url{https://smear.avaa.csc.fi/}); ERA5 data are available via the Copernicus Climate Data Store.}

\authorcontribution{Z.-S. Liu implemented the modelling, coding, and paper drafting; J. Köykkä, W. Peng, A. Kheder, P. Clusius and M. Boy advised on atmospheric interpretation and paper revision.}

\competinginterests{The authors declare that they have no conflict of interest.}

\begin{acknowledgements}
No funds are involved in this research work.
\end{acknowledgements}

\bibliographystyle{copernicus}
\bibliography{main}

\end{document}